# Explainable Deep Belief Network based Auto encoder using novel Extended Garson Algorithm

Satyam Kumar[1,2] and Vadlamani Ravi [1*]
[1] Centre for Artificial Intelligence and Machine Learning
Institute for Development and Research in Banking Technology (IDRBT),
Castle Hills, Masab Tank, Hyderabad 500057, India
[2] School of Computer and Information Sciences (SCIS),
University of Hyderabad, Hyderabad 500046, India
Satyam@idrbt.ac.in; vravi@idrbt.ac.in

**Abstract**
The most difficult task in machine learning is to interpret trained shallow neural networks. Deep neural networks (DNNs) provide impressive results on a larger number of tasks, but it is generally still unclear how decisions are made by such a trained deep neural network. Providing feature importance is the most important and popular interpretation technique used in shallow and deep neural networks. In this paper, we develop an algorithm extending the idea of Garson Algorithm to explain Deep Belief Network based Auto-encoder (DBNA). It is used to determine the contribution of each input feature in the DBN. It can be used for any kind of neural network with many hidden layers. The effectiveness of this method is tested on both classification and regression datasets taken from literature. Important features identified by this method are compared against those obtained by Wald chi square ($\chi^2$). For 2 out of 4 classification datasets and 2 out of 5 regression datasets, our proposed methodology resulted in the identification of better-quality features leading to statistically more significant results vis-à-vis Wald $\chi^2$.

## 1 Introduction

Input features enter into the neural network and an output feature is predicted without understanding the relationship between input and output and hence no explanation is provided by the network (Ripley, 1996). Even though neural networks have great power of providing highly accurate predictions in many real-world datasets, they are still black box models. It is challenging to trade-off high accuracy with high interpretability of black-box models. It is to be noted that neural networks can learn some hidden patterns from data, which increases the bias and makes it difficult for stakeholders to understand predictions made by the black-box models (Varshney & Alemzadeh, 2017). Based on the ideas of model bias and model variance, (Geman et al., 1992) provide a complete examination of the link between learning and generalisation in neural networks. A predetermined model with a high degree of bias that is less reliant on the data may misrepresent the genuine functional relationship. A model-free or data-driven model, on the other hand, can be overly reliant on the particular data and have a high variance. Furthermore, it becomes essential to understand how the predictions are made by neural network for critical domains such as healthcare (Caruana et al., 2015) and finance (Goodman & Flaxman, 2017). To overcome such problems, explainable artificial intelligence (XAI) models have to be designed, which could provide reasoning to the excellent functioning of many machine learning (ML) models including neural works. Explainable AI is the way of providing justification of the decision made by the black box models in human understandable terms, which ensures trust for the stakeholders who use these models. XAI is a human-agent interaction problem (Miller, 2019). Explainable models must be designed in such a way that they do not compromise on the high prediction accuracy. Fig ure. 1 depicts the trade-off between explainability and prediction accuracy for several learning techniques that are used for building ML models (Gunning & Aha, 2019).

---

[*] Corresponding Author: Tel.: +914023294310; Fax: +914023535157.

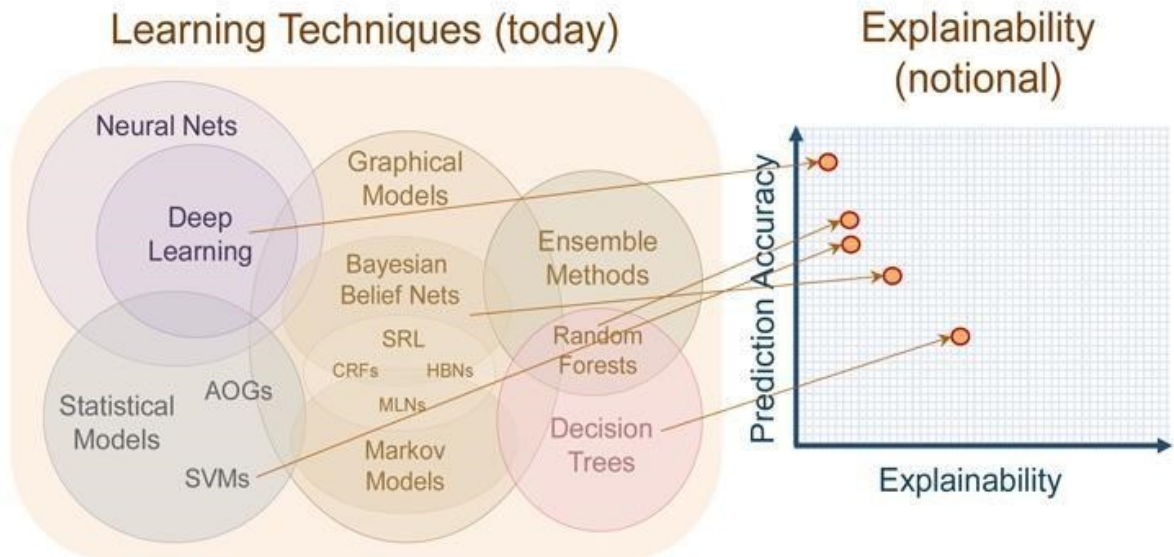

Fig 1. Prediction accuracy versus Explainability for several ML models (Gunning and Aha, 2019).

Variable importance usually addresses the following questions: Does the prediction made by a model depend on a given set of features? (Fisher et al., 2019). Those variables are considered important which does not hamper the performance of the model and hence variable importance is a model-specific explanation technique (Breiman, 2001). As per the European Union's General Data Protection Regulation (GDPR), personal data which is available with business must be able to explain decisions made by the ML models (Goddard, 2017). According to this regulation, affected individuals (for e.g., the user who has been denied loan) has the right to ask the business as to how the system reached that decision.

There are cases in the financial sector where discrimination and biased decisions have been caused by black box models. A study by Freddie Mac found that whites who earn less than $25,000 per year have better credit scores than African Americans who earn between $65,000 and $75,000 per year (Looser, 1999). This resulted in the bank being biased against applicants because 48% of blacks and 27% of whites were thought to have poor credit. According to data obtained from the Home Mortgage Disclosure Act, in 2017 (Dietrich, Liu, Parrish, Roell, & Skhirtladze, 2018)19.3% of blacks were denied loans, only 7.9% of whites and 10.1% of Europeans were denied a loan. A European study found that female entrepreneurs are less likely to access bank loans than male entrepreneurs. It has also been found that male entrepreneurs in Europe are 5% more likely to get a bank loan than female entrepreneurs (Futures, 2019). In recent years, XAI in the financial sector has been applied to solve problems such as financial lending (Chen et al., 2022), financial forecasting (Ilic et al., 2021) and risk management credit (Misheva et al., 2021). Explainable boosted linear regression for next day stock price prediction and feature selection reported by (Carta et.al, 2021)

Deep belief networks (DBN), a class of deep learning neural networks, are probabilistic generative models composed of multi-layer latent (or hidden) binary variables, consisting of stacked Restricted Boltzmann machine (RBM) (Hinton & Salakhutdinov, 2006). The top layers of the DBN consist of the RBM model and the lower layer has a sigmoid belief network. Learning takes place layer by layer with the help of top-down generative weights. This is used to determine the dependence of variables in the previous layer to those in the next layer. Probabilistic learning concepts such as Gibbs Sampling and Contrastive Divergence (CD-k) in training the RBM were introduced by (Hinton, 2002). A fast unsupervised learning algorithm introduced by (Hinton et al., 2006) makes use of greedy layer-by-layer training to learn a deep hierarchical model. The learning procedure requires a single bottom-up pass providing an approximate inference about the values of the top-level hidden variables (Salakhutdinov, 2015). The purpose of using DBN for extracting higher-level dependencies from the input data, which in turn improves the ability of the network, is to capture better representation of the data (Ranzato et al., 2007).

In this paper, extension to the Garson algorithm in the context of DBN-based auto encoder is proposed for identifying feature importance levels in both financial classification datasets and popular regression datasets. In this paper, variables and features are used interchangeably as they are one and the same.

The rest of the paper is organized as follows: Section 2 discusses the literature survey work. Section 3 describes the proposed methodology, namely, the Extended Garson Algorithm. In section 4 the dataset description is provided. Section 5 presents the experimental setup, along with the results obtained for each of the datasets. Finally, section 6 concludes the paper.

## 2 Literature Survey

Estimating input feature contribution levels to the output feature was recognized as an important problem in the early 1990s itself. Garson, (1991) was the first to propose an algorithm that estimates the feature contribution levels to the target feature in a multi-layer perceptron (MLP). Ravi & Zimmermann, (2001) employed Garson algorithm for feature selection while building a fuzzy rule-based classifier. Garson Algorithm was later applied to a shallow network with 1 hidden layer by (Gevrey et al., 2003), (Olden, 2004). Gedeon, (1997) examined the use of weight matrix determining the significance of inputs connected to 2 hidden layers, but it was found to be slow for large networks. Gevrey et al., (2003) compared different methods such as partial derivatives, input perturbation, sensitivity analysis and connection weights (CW) for ecological datasets. Olden, (2004) compared different methodologies on synthetic datasets for assessing the contribution of variables (or features) in multi-layer perceptrons. As synthetic datasets were used, it could not generalize true accuracy and precision as the true importance of the input features is unknown. Garson Algorithm was modified in the context of principal component neural network (PCNN) for bankruptcy prediction by (Ravi & Pramodh, 2008). Later, Garson algorithm was adopted in the context of particle swarm optimization (PSO) trained auto associative neural network (AANN) by (Ravi et al., 2012). Generalization of connection weights for multiple hidden CW for DBN was proposed by (O'Donoghue et al., 2017).

## 3 Proposed Methodology

This section describes the original Garson Algorithm and its proposed extension viz., Extended Garson algorithm (EGA) in the context of DBN-based auto encoder.

### 3.1 Basic Garson Algorithm

Garson, (1991) suggested a method for partitioning the connection weights of a single layer neural network to obtain the relative relevance score for each of the neural network's input variables (features). The black box nature of the neural network is assumed that the connection weights obtained after training the neural network contain more information than would normally be expected (Ravi & Zimmermann, 2001). Garson Algorithm makes use of the absolute values of the connection weights for computing the relative input feature contribution. It does not provide any direction on the relationship between input and output features. Original Garson Algorithm was restricted to only binary classification problems.

Steps of computing relative importance by Garson Algorithm following the notation by (Nath et al., 1997) is as follows:
Let $w_{ij}$ denotes the weight matrix between input-hidden layer where $i^{th}$ neuron of input layer to $j^{th}$ neuron of hidden layer and $w_{jo}$ denotes the weight matrix between hidden layer to output layer, where $j^{th}$ neuron of hidden layer to $o^{th}$ neuron of output layer.

*Step 1: Obtain weight matrix between input-hidden layer ($w_{ij}$) and the weight vector hidden layer-output node ($w_{jo}$)*

*Step 2: Relative contribution is obtained by normalizing the contribution at each hidden neuron by summing the contribution of individual input neurons.*

$$w_{ij}^* = \frac{|w_{ij}|}{\sum_{i=1}^{n} |w_{ij}|} * |w_{jo}|,$$

*where n is the number of input neurons, ∗ represents multiplication*

*Step 3: For each of the input neurons, relative contribution is summed over all hidden neurons and converted into percentage to obtain relative feature importance score.*

## 3.2 Extended Garson Algorithm for DBNA

Relative feature importance in Garson Algorithm is obtained by considering the contribution of each neuron with respect to output neurons. As the number of weights of each hidden layer changes, so does the feature contribution of each neuron and it becomes difficult to keep track of the contribution of those neurons. Increasing the number of hidden layers will change contribution values of neurons in previous layers. Hence, to avoid such difficulties, contribution values of all neurons of particular layers are normalized first, before proceeding to the next hidden layer. All these contributions of hidden layers are stored in a matrix for future use.

Below is the step-by-step methodology of the proposed Extended Garson Algorithm (EGA):

*Step 1. Train Deep Belief Network based Auto-encoder (DBNA) by using Contrastive Divergence (CD) method (Hinton, 2002). Collect all the weights between successive layers. Following the convention of RBM, the output layer of DBNA is considered as the final hidden layer of DBNA.*

*Step 2. Normalization of each of the connection weights matrix of dimension n x m, where n is the number of input neurons and m is hidden neurons is carried out for each of hidden layers.*
*Eq. (1) represents the normalized weight matrix,*

$$\boldsymbol{W} = \begin{bmatrix} \frac{|w_{11}|}{\sum_{i=1}^{n} |w_{i1}|} & \cdots & \frac{|w_{1m}|}{\sum_{i=1}^{n} |w_{im}|} \\ \vdots & \ddots & \vdots \\ \frac{|w_{n1}|}{\sum_{i=1}^{n} |w_{i1}|} & \cdots & \frac{|w_{nm}|}{\sum_{i=1}^{n} |w_{im}|} \end{bmatrix} \quad (1)$$

*Step 3. Normalized weight matrices between two successive layers obtained as in equation (1) are multiplied successively until the final layer is reached to get cumulative weight matrix **CW**, which is of dimension n x n.*

$$\mathbf{CW} = \mathbf{W}^1 . \mathbf{W}^2 . \mathbf{W}^3 \ldots \mathbf{W}^L \quad (2)$$

*where $\boldsymbol{W}^l$ represents the Normalized weight matrix between two successive layers (input and 1st hidden layer) and superscript 1 represents 1st normalized weight matrix. This matrix multiplication is carried till the final layer is reached. L represents the number of hidden layers till the output layer.*

*Step 4. In this step contributions of all hidden layers with respect to each input variable are calculated. All the cumulative weights of each hidden layer are added row wise, to get relative contribution of inputs. The resultant relative contribution (**rc**) vector which of n x 1 and is represented by eq (3)*

$$rc = \begin{bmatrix} \sum_{j=1}^{n} cw_{1j} \\ \sum_{j=1}^{n} cw_{2j} \\ \vdots \\ \sum_{j=1}^{n} cw_{nj} \end{bmatrix} \qquad (3)$$

*Step 5. Relative importance score **(ri)** is the normalized relative contribution for each of the input variables and the range of each normalized value will be in the range of (0, 1). Summation of all values obtained for each of the input variables should be approximately equal to 1. It is of dimension n x 1 and is calculated (in %) to get the contribution of each input variable. Eq (4) represents the **ri** vector*

$$ri = 100\% \cdot \begin{bmatrix} \dfrac{rc_{11}}{\sum_{i=1}^{n} rc_{i1}} \\ \dfrac{rc_{21}}{\sum_{i=1}^{n} rc_{i1}} \\ \vdots \\ \dfrac{rc_{n1}}{\sum_{i=1}^{n} rc_{i1}} \end{bmatrix} \qquad (4)$$

Figure 2 depicts the schematic of proposed methodology, which can be used for solving both classification and regression datasets. Initially, dataset is divided into 80% training data and 20% validation data using stratified random sampling. Training data is used to build DBN and top k features are selected using EGA. Features selected by both Wald $\chi^2$ and EGA are validated on the remaining 20% of data using classification and regression models. The t test is applied using a 10-fold cross validation (cv) on the mean AUC (for classification models) or mean SMAPE (for regression models) values.

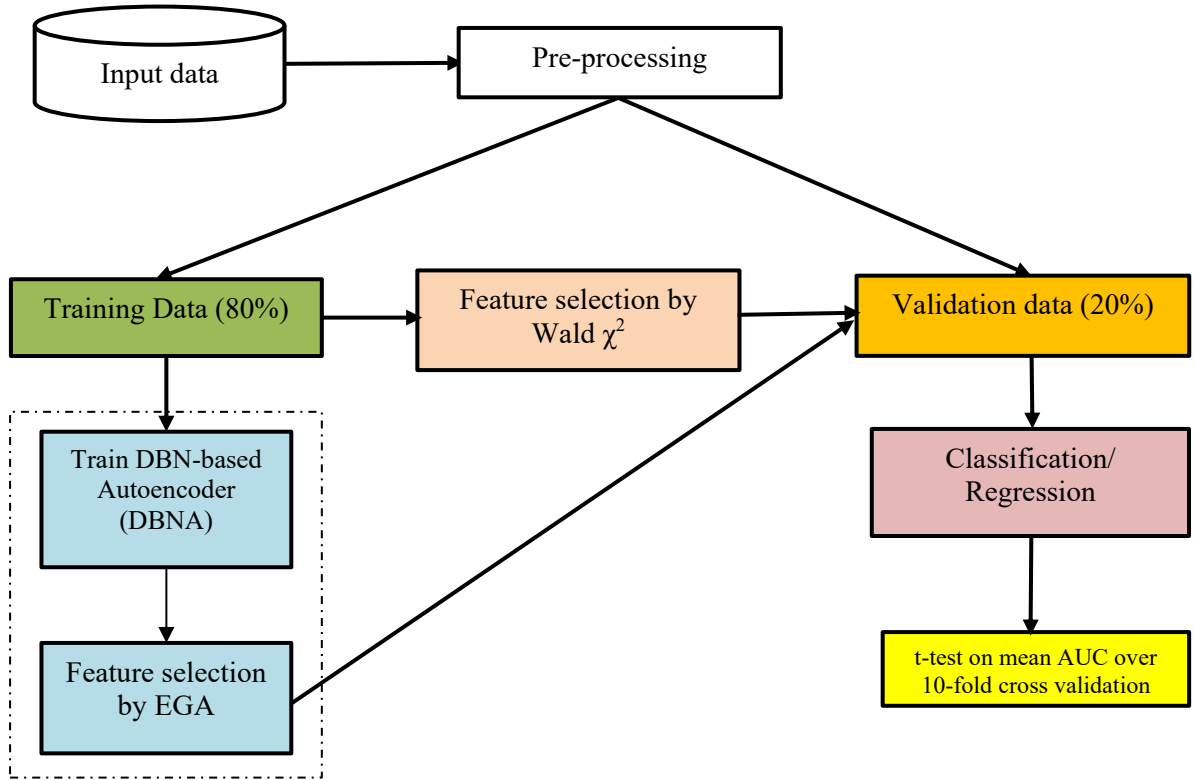

Fig 2. Schematic representation of proposed methodology.

## 4 Description of the datasets

This section describes briefly the classification datasets namely Loan default, Churn Prediction, Credit card fraud detection, Insurance fraud and regression datasets namely Boston Housing, Body Fats, Auto mpg, Forest fires and Pollution.

### 4.1 Classification Datasets

#### 4.1.1 Loan Default

Data is taken from credit card clients located in Taiwan from April 2005 to September (Yeh & Lien, 2009). This dataset has 30000 instances and 24 features, including a target (default payment for next month). Out of 30000 instances, there are 23364 instances that are non-defaults and the remaining 6636 instances are defaulters for next month. Complete details of features provided in Table A.1.

#### 4.1.2 Churn Prediction of customer dataset

Data is taken from a Latin-American bank (Weber, 2004) which has seen a spike in churning of its customers, and they sought to improve its retention strategy. This dataset, obtained from the Business Intelligence Cup held in 2004, has 14814 instances, out of which 13812 are loyal customers and the remaining 1002 are churners, which accounts for 93% and 7% respectively of total customers. The dataset is highly unbalanced. Complete details of features are provided in Table A.2.

#### 4.1.3 Credit card fraud detection

The dataset contains transactions carried by European credit cardholders in September 2013 for 2 days (Dal Pozzolo et al., 2014). Dataset has records of 284807 transactions which out of 492 are fraudulent transactions. This dataset is highly imbalanced with 0.172% of total transactions being fraudulent transactions. This dataset has features $V_1, V_2 \ldots V_{28}$ which are the principal components obtained with principal component analysis (PCA). Time and Amount are the only features for which PCA was not applied. Time represents the number of seconds elapsed from each transaction to the first transaction in the datasets. Feature Class is the target variable, having values of 1 (Fraud) and 0 (Non fraud). For Data balancing, combination of random under sampling and random oversampling was used.

### 4.1.4 Insurance Fraud Detection

Vehicle Insurance fraud dataset (Pyle, 1999) is freely available, provided by Angoss Knowledge Seeker software. This dataset is highly unbalanced, with 94% legal and 6% fraudulent consumers. This dataset is collected in 2 phases. Phase 1 was from January 1994 to December 1995, where 11338 records were collected, and in Phase 2 it was collected from January 1996 to December 1996, where 4083 records were collected. There are 6 numerical and 25 categorical features present in the original dataset, including target variable (fraud or legal). A complete description of each feature is presented in Table A.3.

## 4.2 Regression Datasets

The Boston housing data has 506 instances. Each instance represents about 14 features for homes from various suburbs in Boston, Massachusetts (Harrison & Rubinfeld, 1978). Body fat estimates and various body circumference measurements for 252 men (Behnke & Wilmore, 1974). Automobiles miles per gallon (Auto mpg) data is about fuel consumption measured in miles per gallon. The objective is to predict fuel consumption given 3 multi-valued discrete and 5 continuous attributes and 392 instances (Quinlan, 1993). The forest fires dataset has 517 instances and 13 features. The objective of this dataset is to predict the burned area of forest fires in the northeast region of Portugal, using meteorological and other data. Target (or output) variable area was first transformed with a function $ln(x + 1)$ (Cortez & de Jesus Raimundo Morais, 2007). The pollution dataset includes 9358 instances of hourly averaged responses from an array of 5 metal oxide chemical sensors embedded in an Air Quality Chemical Multisensor Device. The device was located on a field in a significantly polluted area, at road level, within an Italian city. Data collected from March 2004 to February 2005 (1 year duration) (De Vito et al., 2008).

# 5 Result and Discussion

Experiments were carried out on 4 financial binary classification datasets and 5 regression datasets. To train the DBN-based autoencoder, 4 hidden layers were considered to train the model with the learning rate being set to 0.2 and the number of iterations during the model training being set to 100. For assessing the strength of the selected top k features, logistic regression is implemented with solver LIBLINEAR (Fan et al., 2008) and decision tree as implemented in Rokach & Maimon (2014). The area under the curve of Receiver Operating Characteristics Curve (AUC) (Fawcett, 2006) is taken as the performance metric for comparing our proposed methodology with the same classifiers considering all features (i.e., without feature selection).

For regression datasets, 2 hidden layers were considered for training with learning rate set to 0.1 and the number of iterations set to 50. For each of the datasets, standardization of numerical features is carried out. For assessing the strength of the selected top k features linear regression, ridge regression (Hoerl and Kennard, 1970), least absolute shrinkage and selection operator (LASSO) (Tibshirani, 1996), support vector regression (SVR) (Schölkopf & Smola, 2004) and multi-layer perceptron (MLP) (Ruineihart et al., 1986) are employed. Symmetric mean absolute percentage error (SMAPE) (Flores, 1986) is chosen as the performance metric for all the regression datasets except Pollution dataset, where mean absolute percentage error (MAPE) is considered because some original target variable values are indeed zero in this dataset thereby avoiding the division by zero case. Features obtained by the EGA were compared against those obtained by Wald $\chi^2$ for both classification as well as regression datasets. For checking the statistical significance of those features selected by EGA and Wald chi-square method, the t-test is performed under stratified 10-fold cross-validation experimental setup.

Throughout the paper, the following convention is adopted: for 2 hidden layers, the notation (n,m) means that in the DBNA, the 1$^{st}$ hidden layer has n neurons and 2$^{nd}$ hidden layer has m neurons which is the same as the number of input neurons. Similarly for 4 hidden layers, the notation (l,m,n,o) means that in the DBNA, 1st hidden layer has l neurons, 2nd hidden layer has m neurons, 3rd hidden layer has n hidden neurons and last hidden layer has o neurons which is the same as the number of input neurons.

With regard to the classification datasets, the effectiveness of logistic regression and decision tree models with and without feature selection is compared.

As regards the loan dataset, for feature EDUCATION having categorical values such as 0, 5, and 6 is mapped to value 4. Similarly for features from PAY_0 to PAY_6 having categorical values such as 0, -2 is mapped to value -1. Feature such as SEX, EDUCATION, MARRIAGE, PAY_0, PAY_2, PAY_3, PAY_4, PAY_5, PAY_6 one hot encoding is carried out and this approach resulted in a total of 91 features. The models are trained using 91 input features. Synthetic Minority Oversampling TEchnique (SMOTE) (Chawla et al., 2002) is used for data balancing.

As regards the churn prediction dataset, one hot encoding carried out for NCC_T, NCC_T-1, NCC_T-2, N_EDUC, SX, E_CIV features, resulting in 52 features. The model was trained using 52 input features. As regards insurance fraud dataset, it can be observed that the age feature in the data set appears twice, one in numerical form (Age) and the other in categorical form (Age of Policyholder) with numeric values removed from the data to reduce the complexity possessed by multiple unique values (Farquad et al., 2012). Features such as Policy Number, Month, Week of Month, Day of Week, Day of Week Claimed, Week of Month claimed, Year are removed from the dataset. For data balancing, SMOTE is applied.

As regards the credit card fraud detection dataset, features such as Time were removed. Combination of random oversampling followed by random under sampling has been carried out resulting in an accuracy of 55.55% of class 0 and 45.55% of class 1.

As regards Body fat dataset, features like Density were dropped due to high correlation with target variables and thus 12 features were obtained. As regards Auto mpg dataset, one-hot encoding is used for feature origin and obtained features named as country1, country2 and country3. Feature Car Model is dropped. This results in a total of 9 features. As regards Forest fire dataset, Months and days features were one-hot encoded which has resulted in a total of 21 features. As regards pollution dataset, Absolute Humidity is considered as target variable.

For classification and regression data sets, the performance of EGA and Wald $\chi^2$ was tested for statistical significance using the paired t-test performed on the mean AUC (for classification datasets) or mean SMAPE (for regression datasets) at 10+10-2=18 degrees of freedom and 5% level of significance.

Following are the hypothesis for the t-test:
$H_0$: *Features obtained by both EGA and Wald chi-square ($\chi^2$) are statistically the same.*
$H_1$: *Features obtained by either EGA or Wald chi-square ($\chi^2$) are statistically not the same.*

In case of a statistical tie, for classification datasets, EGA or Wald $\chi^2$, whichever yields higher mean AUC with decision tree is chosen as the better method as decision tree is more interpretable compared to logistic regression. Similarly, for regression datasets, EGA or Wald $\chi^2$, whichever yields lowest SMAPE value is chosen as the better method in conjunction with Linear, Ridge and LASSO regression are taken into consideration, as it has better interpretability than SVR and MLP.

## 5.1 Classification Datasets

### 5.1.1 Loan Default

Table 1 presents the results obtained by employing EGA. For 2 hidden layers (75, 91) hidden neurons were considered and for 4 hidden layers (75, 60, 75, 91) hidden neurons were considered. PAY_0_1 turned out to be the most contributing feature which contributes 2.51%. The top 26 features contribute to 54.45% and 55.36% using 2 hidden layers (75, 91) and 4 hidden layers (75, 60, 75, and 91) as depicted in Fig. B.1 and Fig. B.2 respectively, where (75, 91) denotes that the 1st hidden layer has 75 neurons and the 2$^{nd}$ one has 91 neurons. The top 35 features contribute to 68.64% and 69.34% using 2 hidden layers (75,91) and 4 hidden layers (75,60,75,91), as depicted in Fig. B.3 and Fig. B.4 respectively. Table 1 presents that using 26 features yield better AUC score compared to 35 features using both logistic regression and decision trees. While adding 9 additional features increased the feature contribution, it did slightly degrade performance, indicating that the top 26 features were selected as the best to provide explanation to the user.

Table 1. Results of EGA on Loan default data

| Models | Without feature Selection | Top 26 features | | Top 35 features | |
|---|---|---|---|---|---|
| | | (75,91) | (75,60,75,91) | (75,91) | (75,60,75,91) |
| Logistic Regression | 0.6892 | 0.6902 | **0.6912** | 0.6892 | 0.6894 |
| Decision Tree | 0.6130 | **0.6666** | 0.6642 | 0.6625 | 0.6641 |

Results of statistical significance of features identified by EGA and Wald $\chi^2$ for the top 26 features are presented in Table 2. Since the p-values obtained by logistic regression and decision trees value are less than 0.05, the features obtained using Wald $\chi^2$ are significant. Since the mean AUC of the decision tree classifier is higher than that of the logistic regression, the former is preferred.

Table 2. t-test results for Loan default data over 10-fold cross validation

| Models | Mean AUC (EGA) | Mean AUC (Wald $\chi^2$) | t-score | p-value |
|---|---|---|---|---|
| Logistic Regression | 0.5381 | 0.7143 | 107.95 | ≈0 |
| Decision Tree | 0.6694 | **0.7265** | 10.29 | ≈ 0 |

### 5.1.2 Churn Prediction

For 2 hidden layers (40, 52) hidden neurons were considered and for 4 hidden layers, (40, 26, 40, 52) hidden neurons were considered respectively. The SMOTE method is used for data balancing. AGE is the most important feature which contributes 27.85%. The top 8 features contribute upto 84.73% and 98.97% using 2 hidden layers (40, 52) and 4 hidden layers (40, 26, 40, and 52) as depicted in Fig ure. B.5 and Fig ure. B.6 respectively. The top 13 features contribute 84.9% and 98.88% using 2 hidden layers (75,91) and 4 hidden layers (75,60,75,91), as depicted in Fig ure. B.7 and Fig ure. B.8 respectively. Table 3 indicates that 13 features yield better AUC than 8 features using 4 hidden layers (40, 26, 40, and 52) for both logistic regression and decision tree. Adding extra features slightly improves the classification power of both models, suggesting that the top 13 features are likely to provide better explanations.

Table 3. Results of EGA on Churn Prediction data

| Models | Without feature Selection | Top 8 features | | Top 13 features | |
|---|---|---|---|---|---|
| | | (40,52) | (40,26,40,52) | (40,52) | (40,26,40,52) |
| Logistic Regression | **0.8320** | 0.7818 | 0.7818 | 0.7939 | 0.7941 |
| Decision Tree | 0.8097 | 0.7502 | 0.7502 | 0.8059 | **0.8109** |

For statistical significance of features, comparisons between EGA and Wald $\chi^2$ for the top 13 features are presented in Table 4. Since p-values obtained by logistic regression and decision trees re less than 0.05, the features obtained using both EGA and Wald $\chi^2$ are very important. To better understand the significance between EGA and Wald $\chi^2$, the highest mean AUC is considered. The mean AUC of EGA for the decision tree is higher than that of Wald $\chi^2$, so the features obtained by EGA are statistically significant than Wald $\chi^2$. Since the mean AUC of the decision tree classifier with EGA is higher than that of the logistic regression, the former is preferred.

Table 4. t - test results for Churn prediction data over 10-fold cross validation

| Models | Mean AUC (EGA) | Mean AUC (Wald $\chi^2$) | t-score | p-value |
|---|---|---|---|---|
| Logistic Regression | 0.8446 | 0.8446 | 24.7548 | 0 |
| Decision Tree | **0.8973** | 0.784 | 15.819 | $\approx 0$ |

5.1.3 Insurance Fraud

VehicleCategory_1 is the most contributing feature which contributes to 7.93% using 2 hidden layers having following neurons (20, 13). Top 30 features using 2 hidden layers and 4 hidden layers contributes to 70.6278% and 69.2964%, as depicted in Figure. B.9 and Figure. B.10 respectively. Top 45 contributing features using 2 hidden layers and 4 hidden layers are 81.3563% and 80.4442%, as depicted in Figure. B.10 and Figure.B.11 respectively. Table 5 presents that using 30 features results in better AUC score than compared with 45 features using both logistic regression and decision tree. Overall, the top 30 and top 45 features lead to a better AUC than all selected features. With a smaller number of features better explainability is achieved. With fewer features, better interpretability is achieved. While adding 15 more features increases feature contribution, it shows a slight decrease in performance, suggesting that the top 30 selected features are best for providing an explanation to the end user.

Table 5. Results of EGA on Insurance fraud data

| Models | Without feature Selection | Top 30 features | | Top 45 features | |
|---|---|---|---|---|---|
| | | (100,124) | (100,80,100,124) | (100,124) | (100,80,100,124) |
| Logistic Regression | 0.7423 | 0.7581 | **0.7615** | 0.7508 | 0.7550 |
| Decision Tree | 0.5627 | **0.6698** | 0.6621 | 0.5826 | 0.5738 |

For statistical feature significance comparison between EGA and Wald $\chi^2$ for top 30 features is presented in Table 6. Since the p-value obtained by Logistic regression and Decision tree are less than 0.05 and hence features obtained by using both EGA and Wald $\chi^2$ are significant. Mean AUC of Wald $\chi^2$ for decision trees is more than that of EGA, hence features obtained by Wald $\chi^2$ are statistically significant than EGA.

Table 6. t-test results for Insurance fraud data over 10-fold cross validation

| Models | Mean AUC (EGA) | Mean AUC (Wald $\chi^2$) | t-score | p-value |
|---|---|---|---|---|
| Logistic Regression | 0.8236 | 0.8348 | 8.4443 | 9.11 e-07 |
| Decision Tree | 0.9098 | **0.9404** | 23.0249 | 0 |

5.1.4 Credit card fraud

V11 is the most contributing feature, contributing to 20.11% and 20.09% using 2 hidden layers having following neurons (20, 13) and 4 hidden layers having the following neurons (20,14,20,29) respectively. Top 5 features using 2 hidden layers (20, 29) and 4 hidden layers (20, 14, 20, 29) contribute 71.37% and 71.36%, as depicted in Figure. B.13 and Figure. B.14 respectively. Top 14 contributing features using (20, 29) and 4 hidden layers (20, 14, 20,29) are 91.09% and 90.92%, as depicted in Figure. B.15 and Figure. B.16 respectively. The features namely *normAmount* and *V10* are the only different features which have been depicted in Figure. B.14 and Figure. B.16 for the top 5 and 14 contributing features, using 4 hidden layers respectively. Table 7 presents that using 14 features for AUC scores is better than using 5 features using both logistic regression and decision trees. Adding additional features improved the AUC scores of both classification models. The top 14 features selected using logistic regression are used to provide explanations to the end users.

Table 7. Results of EGA on Credit card fraud data

| Models | Without feature Selection | Top 5 features | | Top 14 features | |
|---|---|---|---|---|---|
| | | (20,29) | (20,14,20,29) | (20,29) | (20,14,20,29) |
| Logistic Regression | 0.9441 | 0.9265 | 0.9268 | 0.9485 | **0.9495** |
| Decision Tree | **0.8875** | 0.7904 | 0.7904 | 0.8721 | 0.8567 |

For the statistical comparison of feature importance between EGA and Wald $\chi^2$ for the top 8 features are presented in Table 8. Since p-values were obtained by logistic regression and decision trees lower than 0.05, the features obtained using both EGA and Wald $\chi^2$ are significant. The features obtained by EGA are statistically significant than those obtained by Wald $\chi^2$.

Table 8. t test results for Credit card fraud data over 10-fold cross validation

| Models | Mean AUC (EGA) | Mean AUC (Wald $\chi^2$) | t-score | p-value |
|---|---|---|---|---|
| Logistic Regression | 0.9172 | 0.943 | 42.47 | 0.0004 |
| Decision Tree | **0.986** | 0.98 | 3.377 | 0.0033 |

## 5.2 Regression Datasets

### 5.2.1 Boston Housing

The feature *indus* is the most contributing feature contributing 15.83% using 2 hidden layers. The top 5 and top 9 contribute to 68.61% and 96.47% using 2 hidden layers (20, 13), as depicted in Fig ure. B.17 and Fig ure. B.18 respectively. Table 9 presents that the top 9 features have better SMAPE values for LASSO, SVR and MLP than the same model using the top 5 features. Overall, Ridge regression performs better because it has the lowest SMPE value with the top 5 features selected. Top 5 features that should be offered to users.

Table 9. Results of EGA on Boston Housing data

| Models | Without feature Selection | Top 5 features | Top 9 features |
|---|---|---|---|
| | | (20,13) | (20,13) |
| Linear Regression | 0.9441 | 0.9265 | 0.9485 |
| Ridge | 0.8875 | **0.7904** | 0.8721 |
| LASSO | 21.6678 | 28.9942 | 28.8294 |
| SVR | 12.9764 | 19.0992 | 12.0060 |
| MLP | 15.6203 | 25.9571 | 16.2068 |

To compare feature significance statistics between EGA and Wald $\chi^2$ for the top 9 features is presented in Table 10. Since p-value is greater than 0.05 for all the models, hence features obtained by either EGA or Wald $\chi^2$ are not significant. In this case, the lowest mean SMAPE is taken into account, the features obtained using Ridge regression for Wald $\chi^2$ are more significant than the rest of the models. Therefore, the features obtained using Wald $\chi^2$ are statistically significant than those obtained with EGA.

Table 10. t-test results for Boston Housing data over 10-fold cross validation

| Models | Mean SMAPE (EGA) | Mean SMAPE (Wald $\chi^2$) | t score | p-value |
|---|---|---|---|---|
| Linear Regression | 22.3239 | 21.7576 | 0.155 | 0.878 |
| Ridge | 22.0804 | **21.7412** | 0.0941 | 0.9261 |
| LASSO | 28.5354 | 25.0913 | 0.9006 | 0.3797 |
| SVR | 23.2267 | 23.471 | 0.0701 | 0.9448 |
| MLP | 56.6981 | 72.0737 | 0.8198 | 0.4231 |

### 5.2.2 Body Fat

Weight is the most contributing feature contributing 10.33%. Top 6 and top 10 contribute 57.96% and 91.39% using 2 hidden layers (10.12), as depicted in the Fig ure. B.19 and Fig ure. B.20 respectively. Table 11 presents the contribution of the top 10 features with better SMAPE values for Linear, Ridge and SVR compared to all selected features. The top 6 contributing features are better for MLP than all selected features. Overall, Linear regression has performed better as it has the lowest SMAPE value among all models. Selecting more features not only improves model performance but also better capability to interpret models. Top 10 features must be provided to end users for getting an explanation of the model`s prediction.

Table 11. Results of EGA on Body Fat data

| Models | Without feature Selection | Top 6 features (10,12) | Top 10 features (10,12) |
|---|---|---|---|
| Linear Regression | 20.4945 | 21.3681 | **19.9675** |
| Ridge | 20.5292 | 21.4043 | 20.0358 |
| LASSO | 22.8109 | 22.8109 | 22.8109 |
| SVR | 24.8282 | 25.0257 | 24.1992 |
| MLP | 35.3858 | 30.3665 | 34.7846 |

To compare the feature significance statistics between EGA and Wald $\chi^2$ for the top 6 features is presented in Table 12. Since p-value is greater than 0.05 for all the models, hence features obtained by either EGA or Wald $\chi^2$ are not significant. In this case, the lowest mean SMAPE is taken into account, the features obtained using Ridge regression for Wald $\chi^2$ are more significant than the rest of the models. Therefore, the features obtained using Wald $\chi^2$ are statistically significant than those obtained with EGA.

Table 12. t-test results for Body fat data over 10-fold cross validation

| Models | Mean SMAPE (EGA) | Mean SMAPE (Wald $\chi^2$) | t score | p-value |
|---|---|---|---|---|
| Linear Regression | 22.7185 | 22.3363 | 0.1012 | 0.9205 |
| Ridge | 22.7303 | **22.3111** | 0.111 | 0.9128 |
| LASSO | 25.3124 | 25.3118 | 0.0001 | ≈1 |
| SVR | 32.148 | 32.3396 | 0.0371 | 0.971 |
| MLP | 44.9821 | 43.6147 | 0.1704 | 0.8666 |

### 5.2.3 Auto mpg

The feature *cylinders* is the most contributing feature contributing 25.30%. Top 3 and top 5 contribute 73.97% and 96.91% using 2 hidden layers (13.7), as shown in the Fig ure. B.21 and Fig ure. B.22 respectively. Table 13 presents that there is no better regression model for the selected top k features (k = 3.5). Overall, SVR performs better because it has the lowest SMAPE value, suggesting that selecting the top 5 features has greater explanatory power because the SMAPE of the top 5 selected features is closer to the value of all selected features.

Table 13. Results of EGA on Auto mpg data

| Models | Without feature Selection | Top 3 features (13,7) | Top 5 features (13,7) |
|---|---|---|---|
| Linear Regression | 11.6920 | 15.1716 | 14.7492 |
| Ridge | 11.6986 | 15.1274 | 14.7430 |
| LASSO | 18.1792 | 26.1866 | 22.1011 |
| SVR | **9.1420** | 12.3489 | 11.6627 |
| MLP | 10.0705 | 16.7948 | 13.5329 |

To compare feature significance statistics between EGA and Wald $\chi^2$ for the top 5 features are presented in Table 14. As p-values are less than 0.05 for all models, except for linear regression, the features obtained by both EGA and Wald $\chi^2$ are significant. In this case, the mean SMAPE is the lowest among the Ridge, LASSO, SVR and MLP considered. Features obtained using SVR for EGA are more significant than the rest of the models. Therefore, the features obtained by EGA are statistically significant than Wald $\chi^2$.

Table 14. t-test results for Auto mpg data over 10-fold cross validation

| Models | Mean SMAPE (EGA) | Mean SMAPE (Wald $\chi^2$) | t score | p-value |
|---|---|---|---|---|
| Linear Regression | 12.8217 | 21.878 | 5.887 | 0.1138 |
| Ridge | 12.8249 | 21.825 | 5.674 | 0.0004 |
| LASSO | 26.1711 | 25.76 | $\approx 0$ | $\approx 0$ |
| SVR | **11.684** | 21.514 | 4.291 | $\approx 0$ |
| MLP | 112.437 | 28.124 | 4.563 | 0.0002 |

5.2.4 Forest fires

DC is the most contributing feature, contributing 16.4219%. Top 8 and top 15 contribute 78.24% and 93.73% using 2 hidden layers (15.21), as depicted in the Fig ure. B.23 and Fig ure. B.24 respectively. Table 15 presents the contributions of the top 15 features with better SMAPE values for Linear, Ridge and MLP compared to all selected features. Overall, Ridge regression performs better because it has the lowest SMAP value among all regression models, suggesting that choosing the top 15 features has greater explanatory power. Adding more features that improved model performance for Linear Regression, Ridge Regression, and LASSO Regression on selected features.

Table 15. Results of EGA on Forest fires data

| Models | Without feature Selection | Top 8 features (15,21) | Top 15 features (15,21) |
|---|---|---|---|
| Linear Regression | 128.2780 | 130.9742 | 127.7935 |
| Ridge | 128.2404 | 130.9782 | **127.7613** |
| LASSO | 132.7309 | 132.7309 | 132.7309 |
| SVR | 136.3574 | 147.3458 | 137.8894 |
| MLP | 131.4590 | 133.5095 | 130.9400 |

To compare feature significance statistics between EGA and Wald $\chi^2$ for the top 8 features are presented in Table 16. Since p-values are greater than 0.05 for all models, the features obtained by EGA or Wald $\chi^2$ are therefore not significant. In this case, the mean SMAPE is the lowest among all the models taken into account, the features obtained using Ridge regression for EGA are more significant than the rest of the models. Therefore, the features obtained by EGA are statistically significant than Wald $\chi^2$.

Table 16. t-test results for Forest fires data over 10-fold cross validation

| Models | Mean SMAPE (EGA) | Mean SMAPE (Wald $\chi^2$) | t score | p-value |
|---|---|---|---|---|
| Linear Regression | 130.757 | 137.766 | 0.8616 | 0.4 |
| Ridge | **129.986** | 137.5984 | 0.940 | 0.359 |
| LASSO | 136.938 | 136.616 | 0.051 | 0.959 |
| SVR | 161.0375 | 161.1277 | 0.022 | 0.982 |
| MLP | 179.941 | 171.9102 | 1.4954 | 0.1521 |

### 5.2.5 Pollution

PTO8_S2_NHMC is the largest contributor, accounting for 14.1927%. Using two hidden layers (15, 10), as shown in Figures B.25 and B.26, the top 5 and top 7 contribute 67.13% and 87.53%, respectively.

Table 17 presents a comparison of the MAPE values of the top k (k = 3.5) selected features and in which the feature is not selected. None of the selected contributing features is dominant in the explanation, Even though PTO8_S2_NMHC is presented as the most important and contributing feature, but none of the selected features are suitable for explanation to end users. Overall, SVR has the lowest MAPE value. This shows that the selection of the top 7 features is more explanatory because the MAPEs of the top 7 selected features are close to the MAPEs of all the selected features.

Table 17. Results of EGA on Pollution data

| Models | Without feature Selection | Top 5 features (15,10) | Top 7 features (15,10) |
|---|---|---|---|
| Linear regression | 0.1456 | 0.3015 | 0.2897 |
| Ridge | 0.1456 | 0.3015 | 0.2897 |
| LASSO | 0.4084 | 0.4101 | 0.4084 |
| SVR | **0.0889** | 0.3001 | 0.2233 |
| MLP | 0.0951 | 0.3281 | 0.2454 |

For statistical feature significance comparison between EGA and Wald $\chi^2$ for top 5 features is presented in Table 18. The features obtained by either EGA or Wald $\chi^2$ are not important in SVR and MLP, as all models except SVR and MLP have p-values less than 0.05. SVR and MLP are difficult to explain and can be ignored. In this case, the lowest mean MAPE of Linear, Ridge, and LASSO is considered among the models in which the features obtained in both EGA and Wald $\chi^2$ are important. Linear and ridge regressions have the lowest MAPE values for Wald $\chi^2$, so the features obtained with Wald $\chi^2$ are statistically more significant than EGA.

Table 18. t-test results for Pollution data over 10-fold cross validation

| Models | Mean MAPE (EGA) | Mean MAPE (Wald $\chi^2$) | t score | p-value |
|---|---|---|---|---|
| Linear Regression | 0.574 | **0.4551** | 10.384 | ≈ 0 |
| Ridge | 0.5742 | **0.4551** | 10.384 | ≈ 0 |
| LASSO | 0.7977 | 0.6940 | 5.8481 | ≈ 0 |
| SVR | 0.5994 | 0.6 | 0.061 | 0.952 |
| MLP | 2.1861 | 1.82 | 1.196 | 0.247 |

Adding a hidden layer does not improve the relative feature score because the DBN-based autoencoder becomes saturated with training. Using 4 hidden layers, it was identified that the same features were obtained as the 2 hidden layers. The feature order is changed and a hidden layer with a slightly modified feature contribution score is added, but the AUC score of the classification dataset is not affected.

With regard to classification datasets, datasets with a higher number of categorical variables, such as loan default and insurance fraud, will require a lesser number of features to be picked for interpretability. Both approaches are equally interpretable for the specified features. Higher numbers of features are needed for greater interpretability when the number of numerical values in datasets rises, as is the case with the Churn Prediction and Credit Card Fraud Detection datasets. For those selected features, either approach results in interpretability, indicating that increased numeric values for interpretability affect DBNA.

With regard to regression datasets, a greater number of features is necessary for DBNA to offer better interpretability, which subsequently enhances model performance. With the exception of the Boston Housing dataset, where fewer features are needed for interpretability. With respect to the datasets for the Boston Housing and Forest Fires, Ridge Regression performed better on the features that were chosen, leading to improved model performance than in the absence of feature selection. Similar results were seen for the Auto mpg and Pollution datasets, wherein SVR without feature selection outperformed SVR with feature selection. In order to improve model performance and interpretability for body fat, linear regression with a greater number of carefully chosen features is needed.

For classification datasets, even with a smaller number of features, an explanation can be given. For the 2 datasets, i.e. churn prediction and credit card fraud detection, EGA identified a statistically significant feature compared to Wald $\chi^2$ and for the other 2 datasets, i.e. data set on loan and insurance fraud, Wald $\chi^2$ determined statistically significant characteristics of the features. For the regression dataset, adding additional features leads to better explanations, but poor model performance for the dataset, as seen in the Pollution and Auto mpg datasets. Therefore, the explanation obtained for the regression data sets becomes less reliable, as the performance of the model is compromised.

# 6 Conclusions

A novel EGA is proposed to understand the behavior of the DBN-based autoencoder which is black box model, by estimating the feature contribution factors and allowing it to be explained by making it scalable to any number of hidden layers. Its effectiveness has been demonstrated using banking and insurance classification datasets and literature benchmark regression datasets. For the classification dataset, the characteristics selected by the EGA were statistically more significant than the characteristics from the Wald $\chi^2$, except for the loan default and insurance fraud datasets. With the exception of the Auto mpg and Forest fires datasets, all other regression datasets provided significant features for Wald $\chi^2$ compared to the proposed methodology.

This methodology can be further extended to the convolutional neural network (CNN) (LeCun & Bengio, 1998). In addition, the Extended Garson algorithm can be compared in the future with a Local Interpretable Model-agnostic Explanation (LIME) (Ribeiro et al., 2016).

Appendix A. Tabular description of Classification datasets

Table A.1. Description of Loan default dataset

| Feature | Description |
|---|---|
| LIMIT_BAL | Amount of given credit in NT dollars (Includes individual and family/supplementary credit |
| SEX | Gender (1=male, 2=female) |
| EDUCATION | 1=graduate school, 2=university,3=high school, 4=others, 5=unknown, 6=unknown |
| MARRIAGE | Marital status (1=married,2=single, 3=others) |
| AGE | Age in years |
| PAY_0 | Repayment status in September, 2005 (-1=pay duly,1=payment delay for one month, 2=payment delay for two months,8=payment delay for eight months, 9=payment delay for nine months and above |
| PAY_2 | Repayment status in August, 2005 (scale same as PAY_0) |
| PAY_3 | Repayment status in July, 2005 (scale same as above) |
| PAY_4 | Repayment status in June, 2005 (scale same as above) |
| PAY_5 | Repayment status in May, 2005 (scale same as above) |
| PAY_6 | Repayment status in April, 2005 (scale same as above) |
| BILL_AMT1 | Amount of bill statement in September, 2005 (NT dollar) |
| BILL_AMT2 | Amount of bill statement in August, 2005 (NT dollar) |
| BILL_AMT3 | Amount of bill statement in July, 2005 (NT dollar) |
| BILL_AMT4 | Amount of bill statement in June, 2005 (NT dollar) |
| BILL_AMT5 | Amount of bill statement in May, 2005 (NT dollar) |
| BILL_AMT6 | Amount of bill statement in April, 2005 (NT dollar) |
| PAY_AMT1 | Amount of previous payment in September, 2005 (NT dollar) |
| PAY_AMT2 | Amount of previous payment in August, 2005 (NT dollar) |
| PAY_AMT3 | Amount of previous payment in July, 2005 (NT dollar) |
| PAY_AMT4 | Amount of previous payment in June, 2005 (NT dollar) |
| PAY_AMT5 | Amount of previous payment in May, 2005 (NT dollar) |
| PAY_AMT6 | Amount of previous payment in April, 2005 (NT dollar) |
| default.payment.next.month | Default payment (1=yes, 0=no) |

Table A.2. Description of Churn prediction dataset

| Features | Description | Value |
|---|---|---|
| Target | Target variable | 0    Non churner (nCh), <br> 1    Churner (Ch) |
| CRED_T | Credit in month T | Positive real number |
| CRED_T-1 | Credit in month T-1 | Positive real number |
| CRED_T-2 | Credit in month T-2 | Positive real number |
| NCC_T | Number of credit cards in month T | Positive integer |
| NCC_T-1 | Number of credit cards in month T-1 | Positive integer |
| NCC_T-2 | Number of credit cards in month T-2 | Positive integer |
| INCOME | Customer's income | Positive real number |
| N_EDUC | Customer's educational level | 1. University student, <br> 2. Medium degree <br> 3. Technical degree, <br> 4. University degree |
| AGE | Customer's age | Positive integer |
| SX | Customers sex | 1 – Male, 0 – Female |
| E_CIV | Civilian status | 1– Single, 2 – Married, 3 - Widow, 4 - Divorced–Widow,4 – Divorced |
| T_WEB_T | Number of web transactions in month T | Positive integer |
| T_WEB_T-1 | Number of web transactions in month T-1 | Positive integer |
| T_WEB_T-2 | Number of web transactions in month T-2 | Positive integer |
| MAR_T | Customer's margin for the company in month T | Real number |
| MAR_T-1 | Customer's margin for the company in month T-1 | Real number |
| MAR_T-2 | Customer's margin for the company in month T-2 | Real number |
| MAR_T-3 | Customer's margin for the company in month T-3 | Real number |
| MAR_T-4 | Customer's margin for the company in month T-4 | Real number |
| MAR_T-5 | Customer's margin for the company in month T-5 | Real number |
| MAR_T-6 | Customer's margin for the company in month T-6 | Real number |

Table A.3. Description of Insurance fraud dataset

| Feature | Value |
|---|---|
| Month | January to December |
| Week of Month | 1,2,3,4,5 |
| Day of week | Sunday to Saturday |
| Month claimed | January to December |
| Week of month claimed | 1, 2, 3, 4, 5 |
| Day of week claimed | Sunday to Saturday |
| Year | 1994, 1995 and 1996 |
| Make | Accura, BMW, Chevrolet, Dodge, Ferrari, Ford, Honda, Jaguar, Lexus, Mazda, Mercedes, Mercury, Nissan, Pontiac, Porsche, Saab, Saturn, Toyota, VW |
| Accident Area | Rural, Urban |
| Sex | Male, Female |
| Marital Status | Divorced, Married, Single, Widow |
| Age | From 16 to 80 |
| Fault | Policyholder, Third Party |
| Policy Type | Sedan—all perils, Sedan—collision, Sedan—liability, Sport—all perils, Sport—collision, Sport—liability, Utility—all perils, Utility—collision, Utility—liability |
| Vehicle Category | Sedan, Sport, Utility |
| Vehicle Price | (Less than $20,000), ($20,000 - $29,000), ($30,000 - $39,000), ($40,000 - $59,000), ($60,000 - $69,000), (greater than $69,000) |
| Policy Number | 1-15420 |
| Rep Number | 1-16 |
| Deductible | 300, 400, 500, 700 |
| Driver Rating | 1, 2, 3, 4 |
| Days Policy Claims | 15 - 30,8 - 15, more than 30, none |
| Days Policy Accident | 1 - 7, 15 - 30, 8 - 15, more than 30, none |
| Past Number of Claims | 1,2 - 4, more than 4, none |
| Age of Vehicle | 2 years, 3 years, 4 years, 5 years,6 years, 7 years, more than 7 years, new |
| Age of Policy Holder | 16 - 17, 18 - 20, 21 - 25, 26 - 30, 31 - 35, 36 - 40, 41 - 50, 51 - 65, over 65 |
| Police Report Filed | Yes, No |
| Witness Present | Yes, No |
| Agent Type | External, Internal |
| Number of Supplements | 1 - 2, 3 - 5, more than 5, none |
| Address Change Claim | Under 6 months, 1 year, 2 - 3 years, 4 - 8 years, no change |
| Number of Cars | 1, 2 3 - 4 5 - 8, more than 8 |
| Base Policy | All perils, Collision, Liability |
| Fraud Found | 0, 1 (response) |

# Appendix B. Figures of EGA feature importance

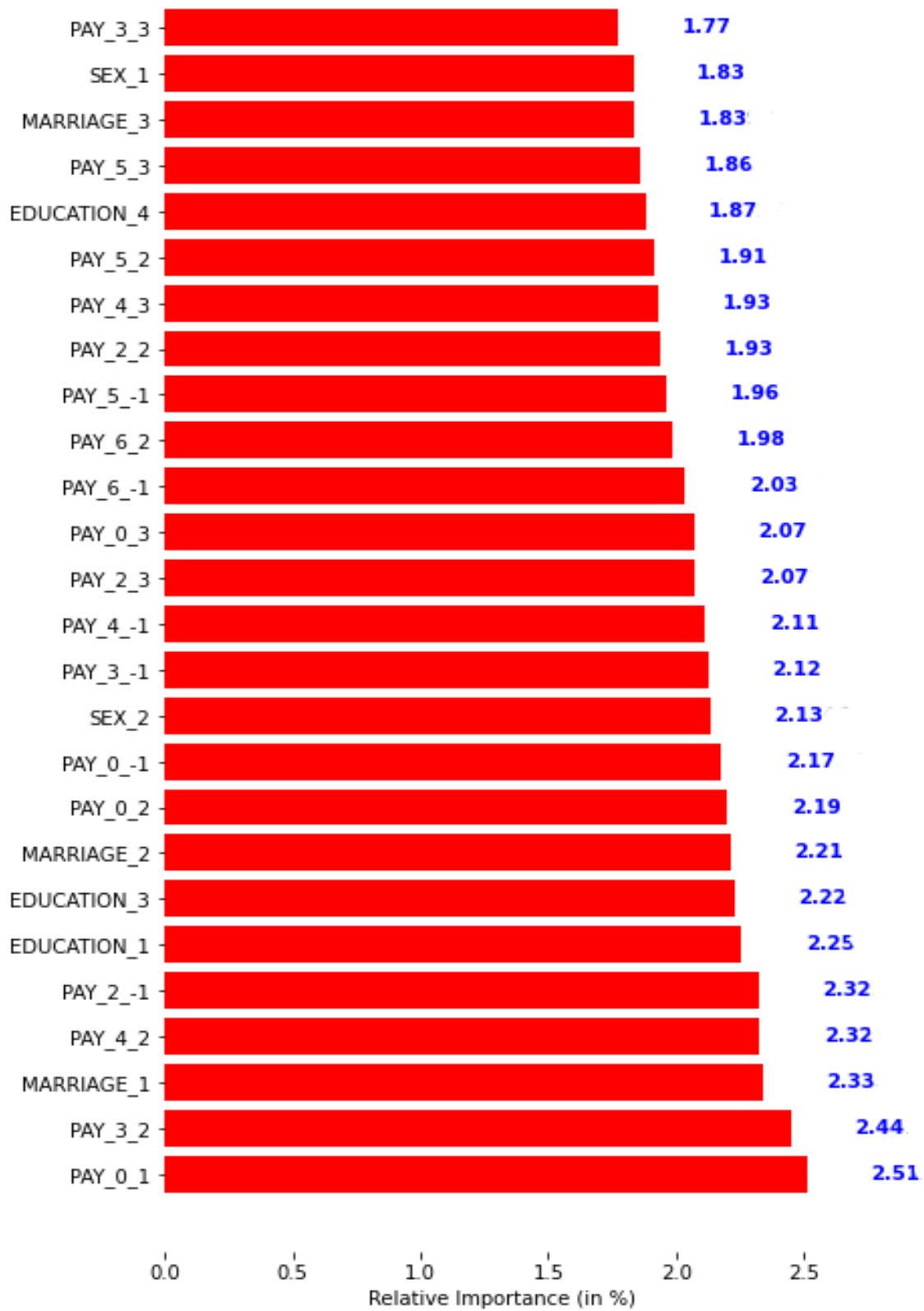

Fig B.1. Contribution of top 26 features using (75, 91) hidden neurons

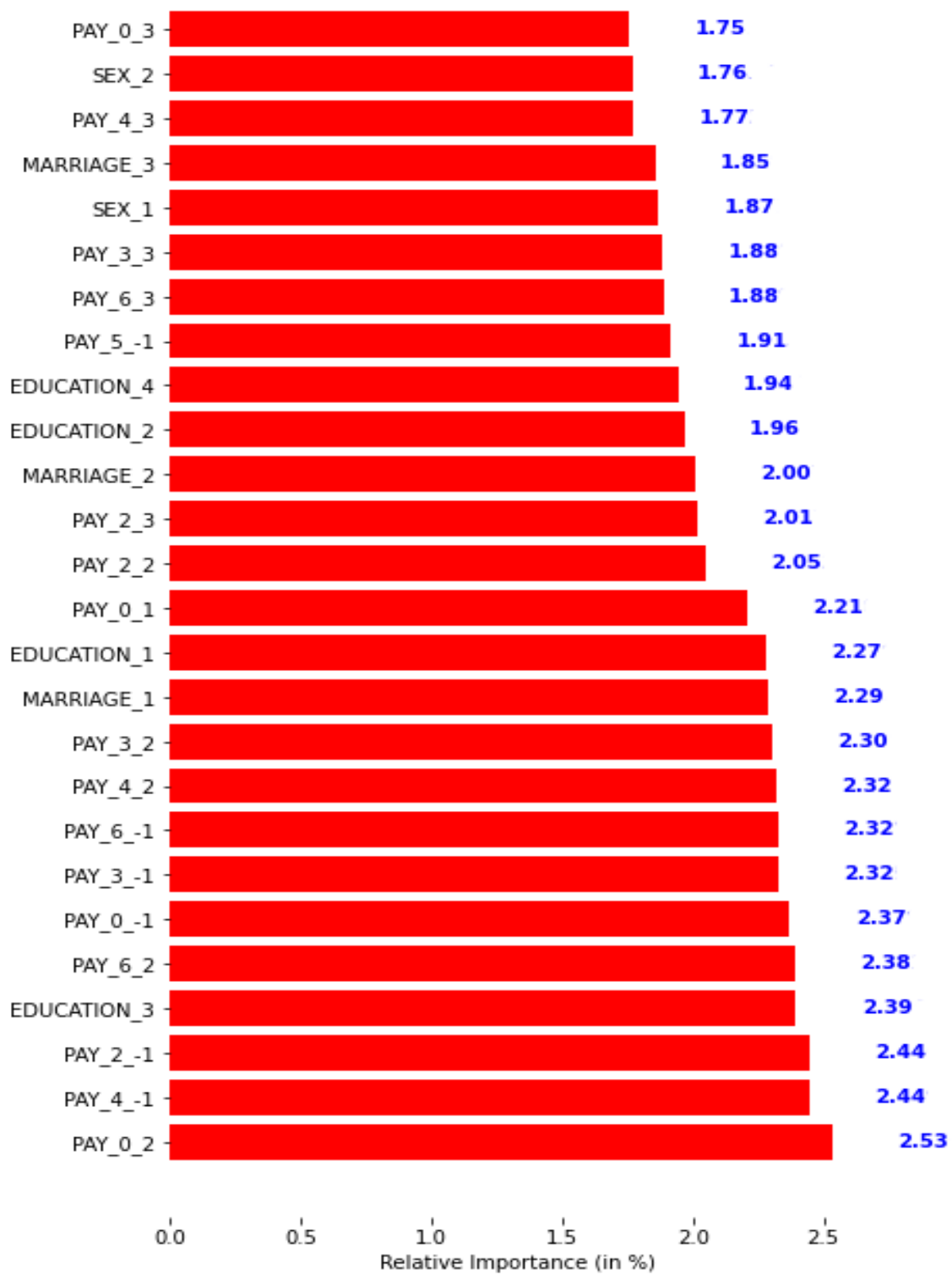

Fig B.2. Contribution of top 26 features using (75, 60, 75, 91) hidden neurons

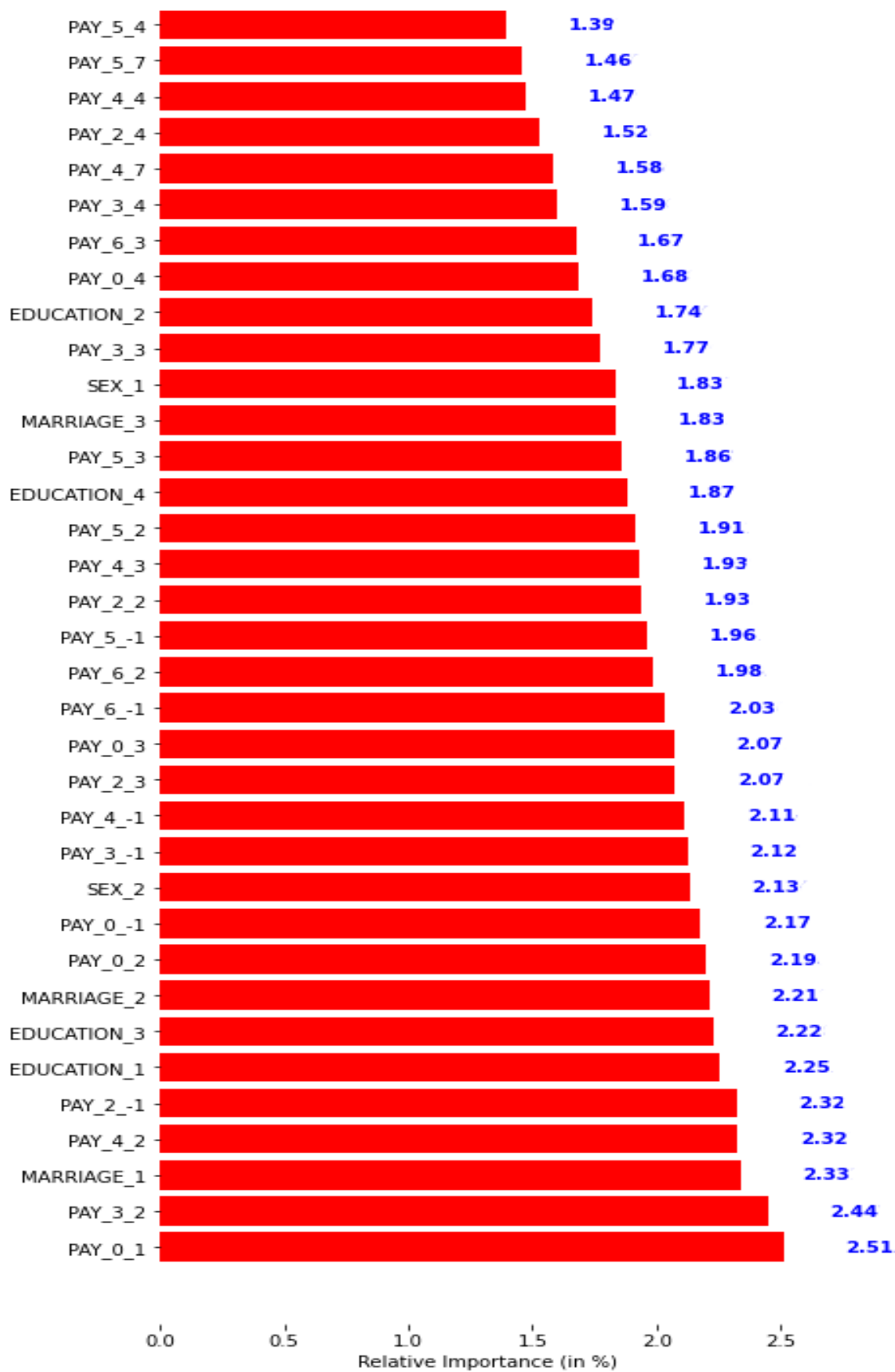

Fig B.3. Contribution of top 35 features using (75, 91) hidden neurons

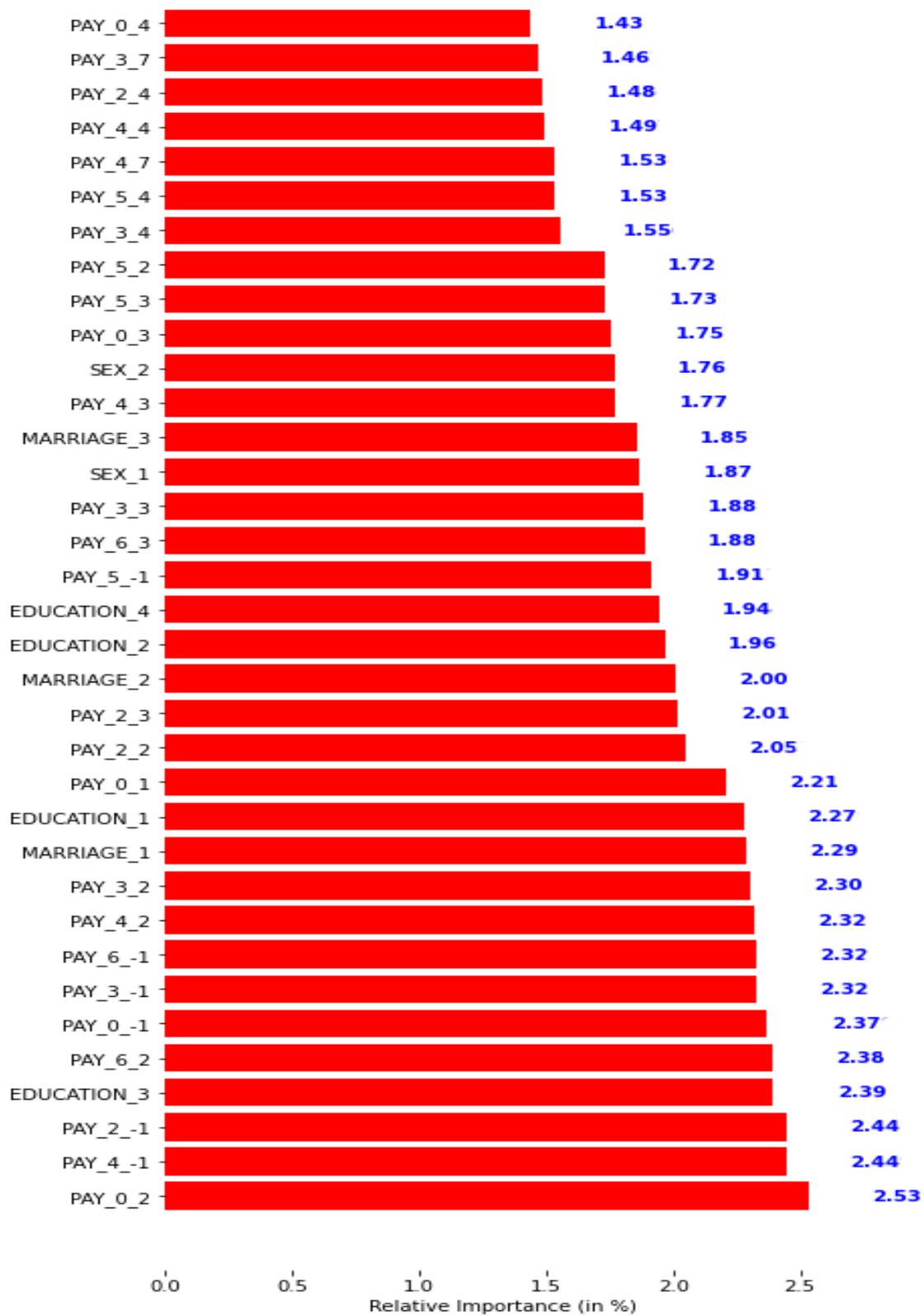

Fig B.4. Contribution of top 35 features using (75, 60, 75, 91) hidden neurons

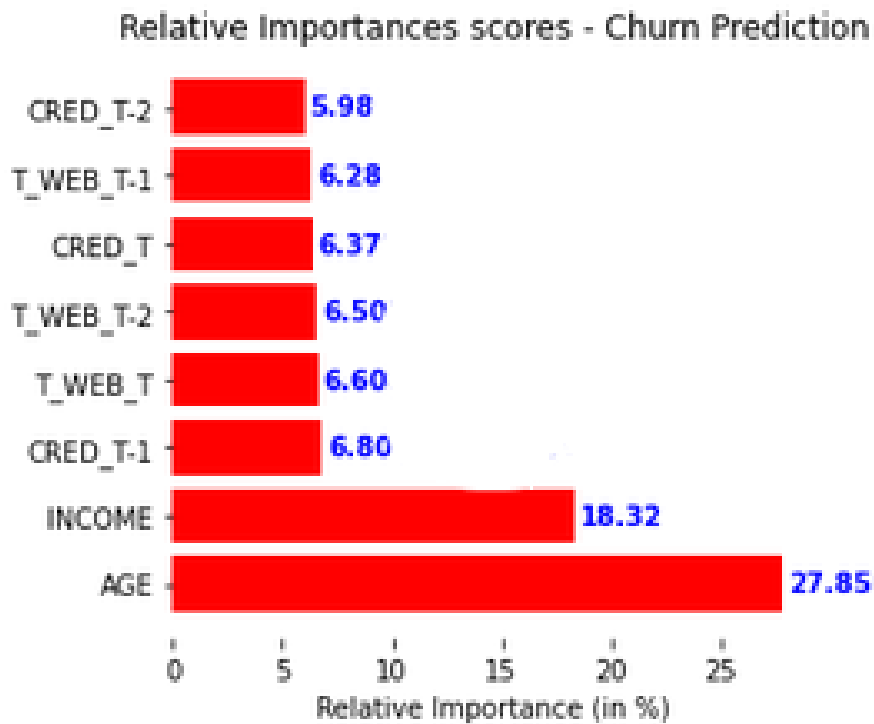

Fig B.5. Contribution of top 8 features using (40, 52) hidden neurons

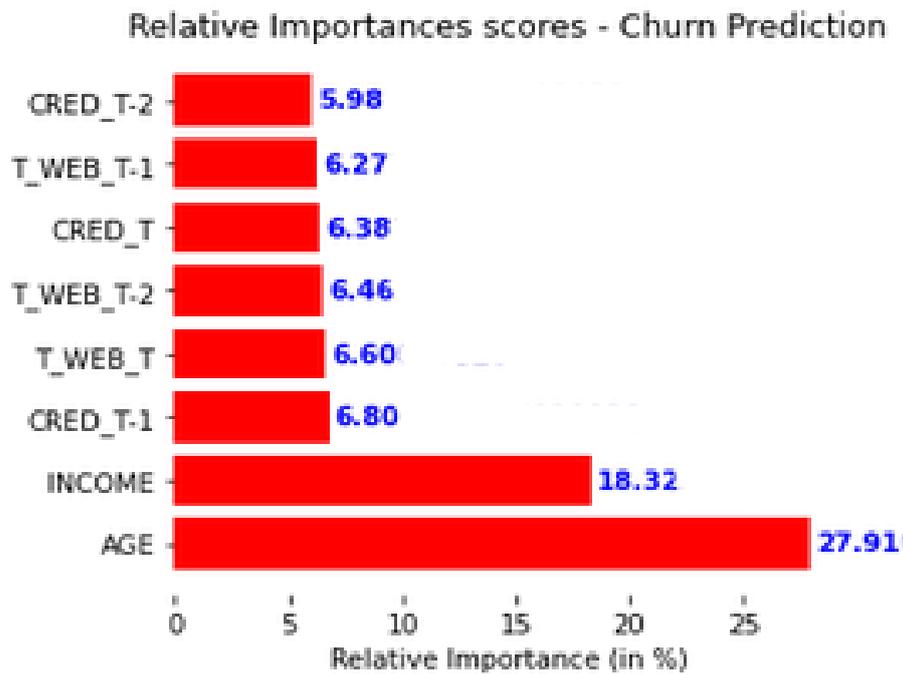

Fig B.6: Contribution of top 8 features using (40, 26, 40, 52) hidden neurons

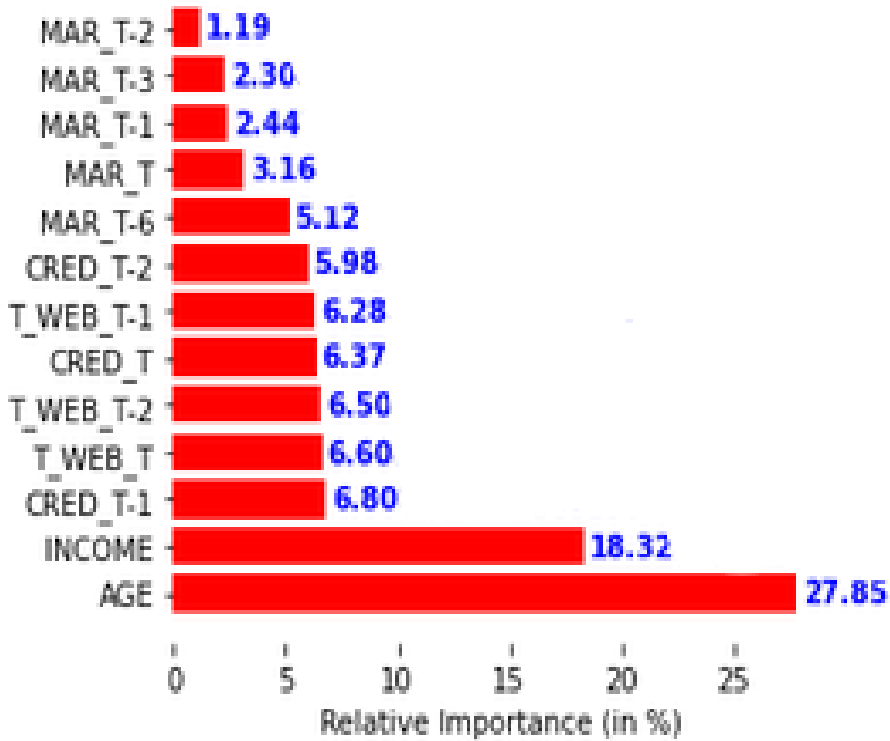

Fig B.7: Contribution of top 13 features using (40, 52) hidden neurons

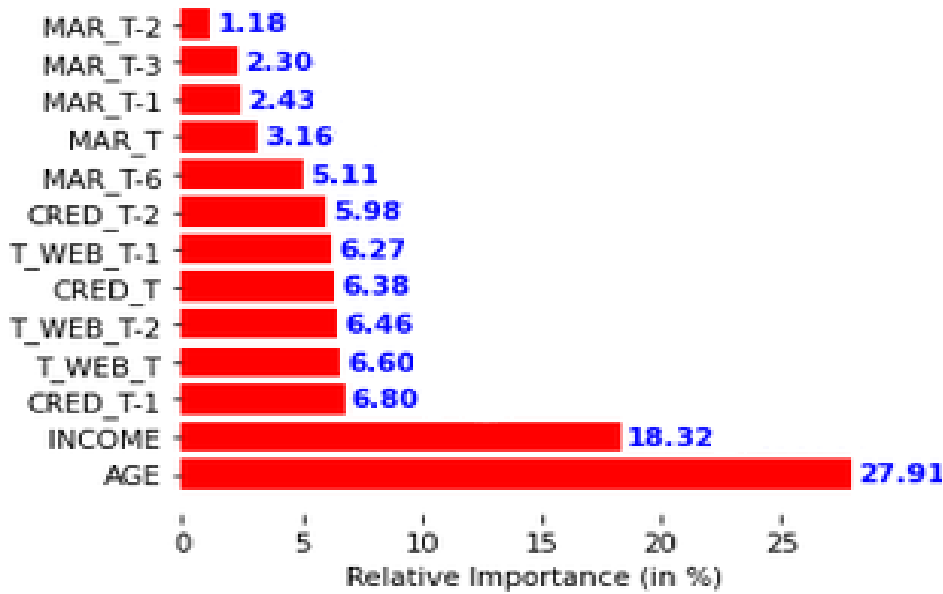

Fig B. 8: Contribution of top 13 features using (40, 26, 40, 52) hidden neurons

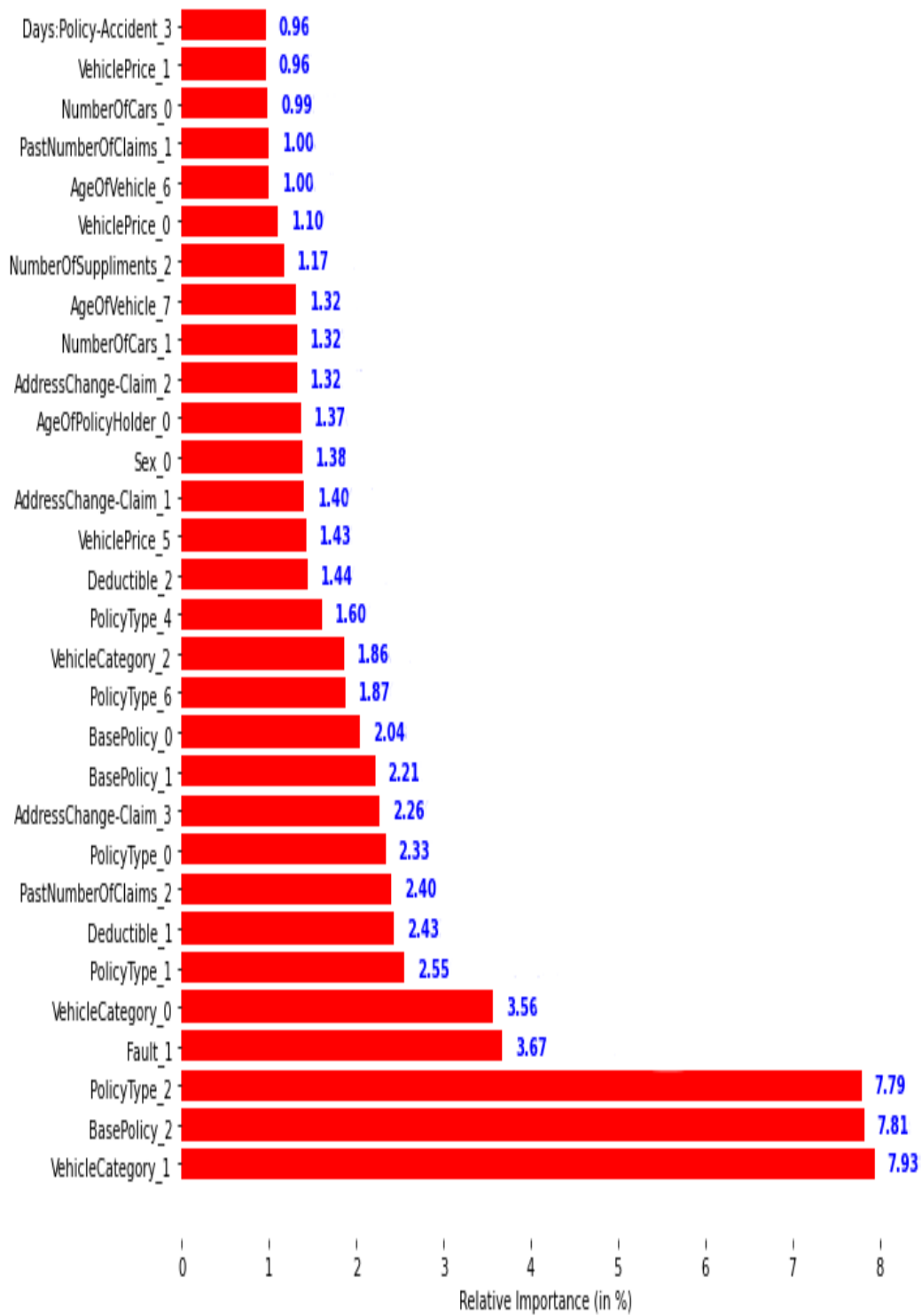

Fig B 9: Contribution of top 30 features using (100,124) hidden neurons

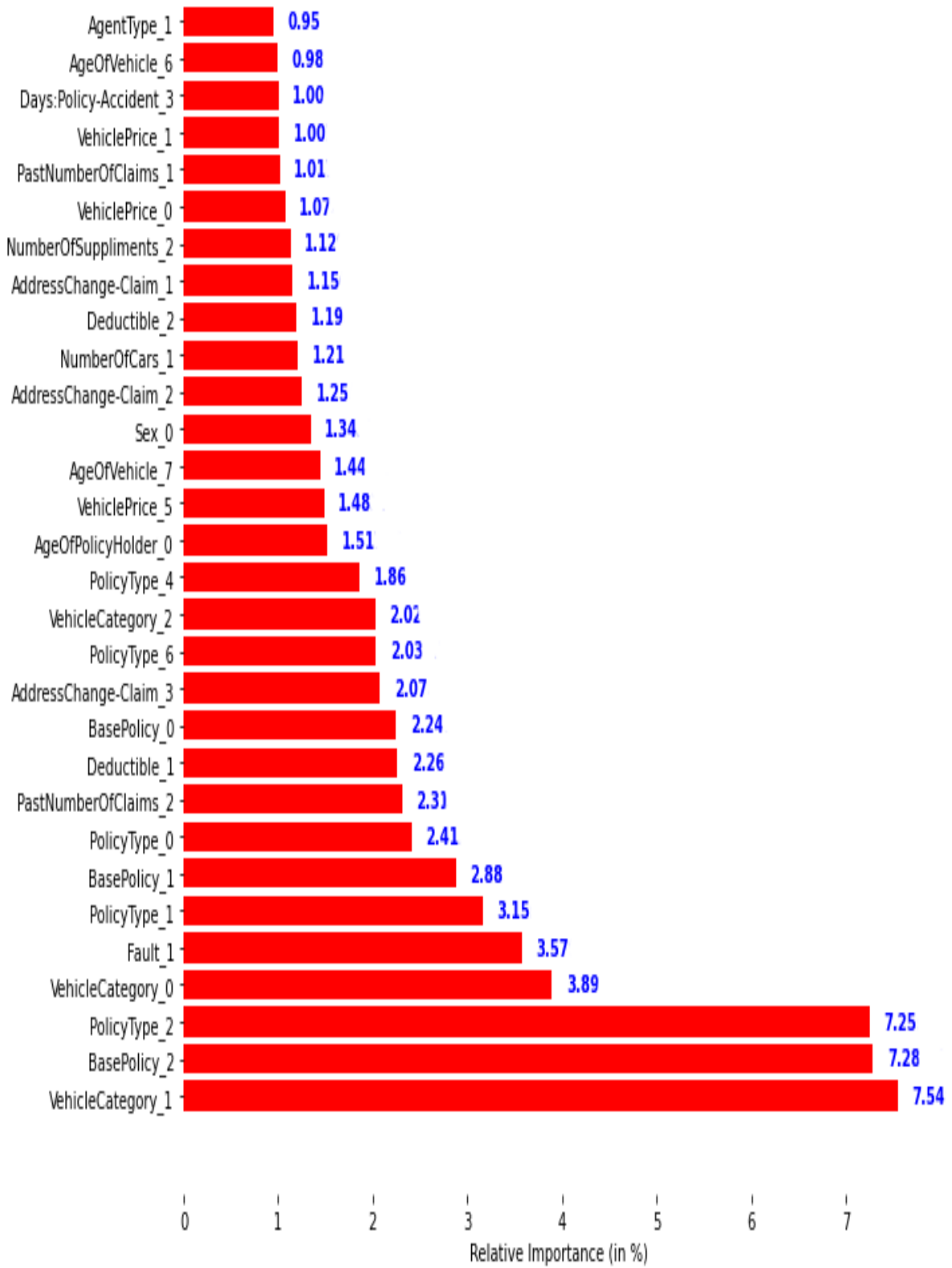

Fig B 10: Contribution of top 30 features using (100, 80,100,124) hidden neurons

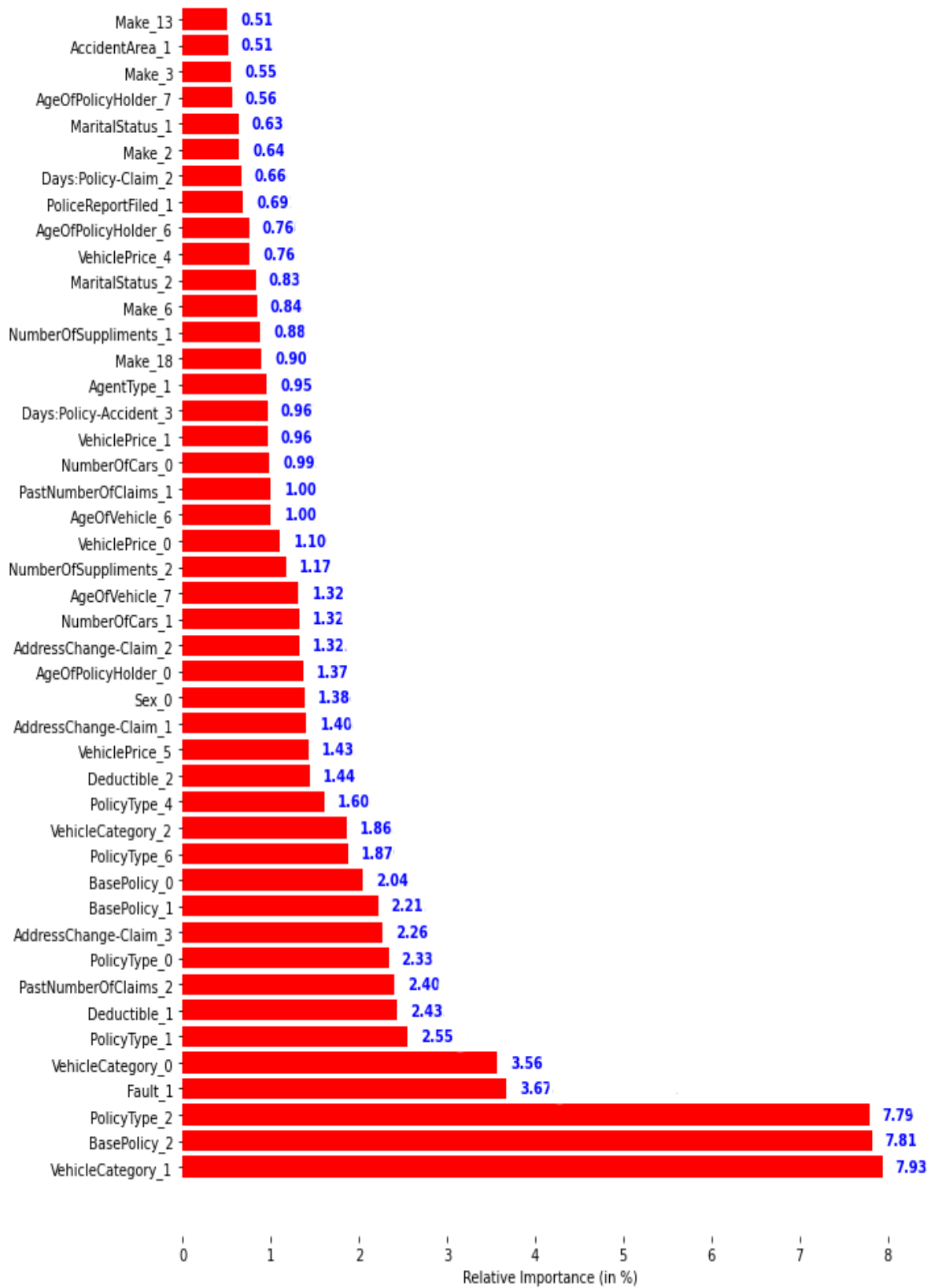

Fig B 11: Contribution of top 45 features using (100,124) hidden neurons

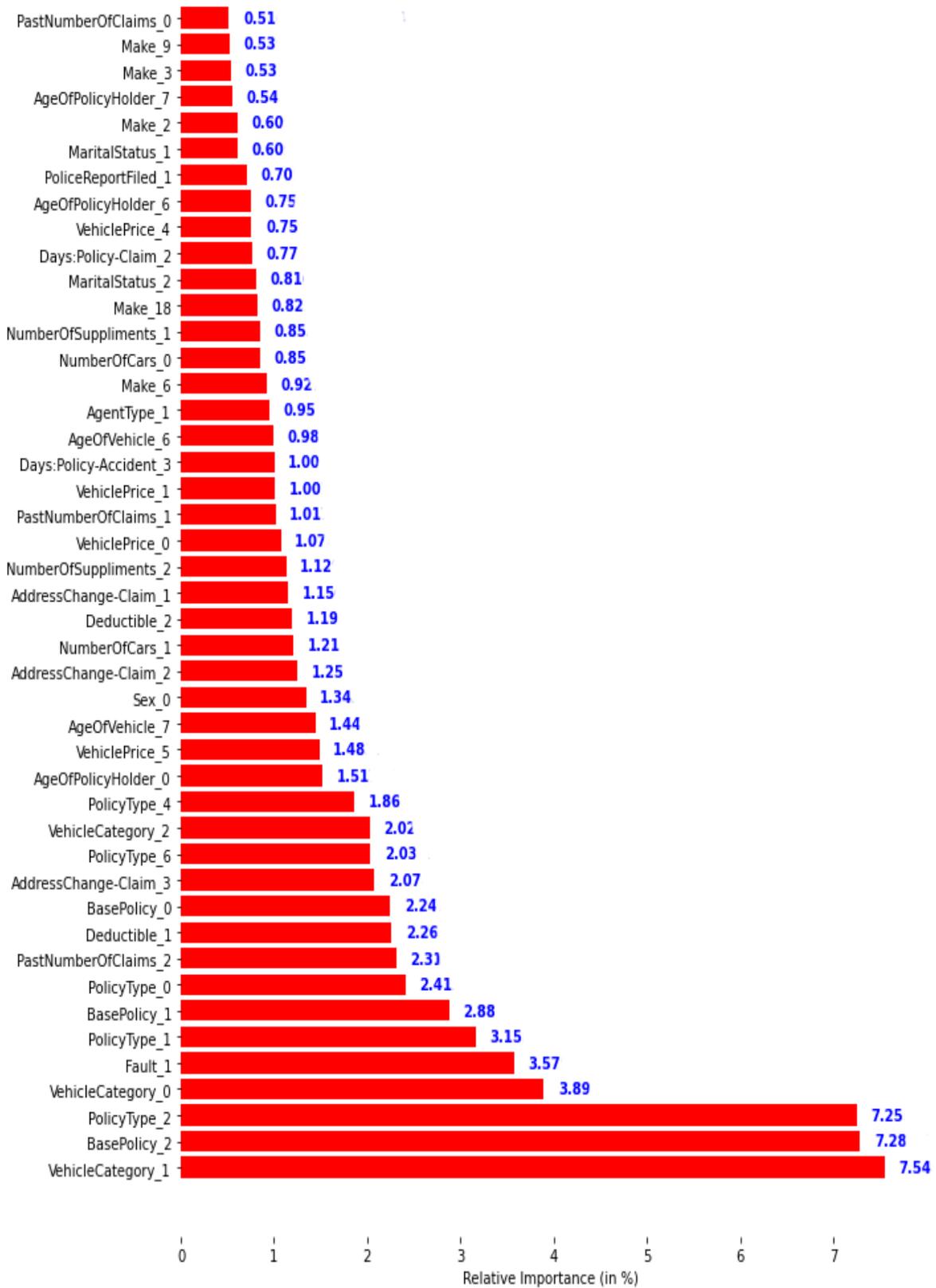

Fig B 12: Contribution of top 45 features using (100, 80,100,124) hidden neurons

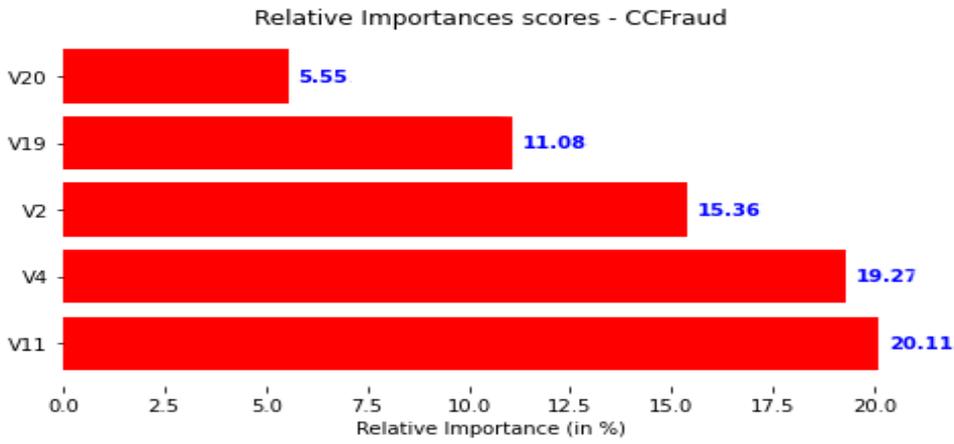

Fig B 13: Contribution of top 5 features using (20, 29) hidden neurons

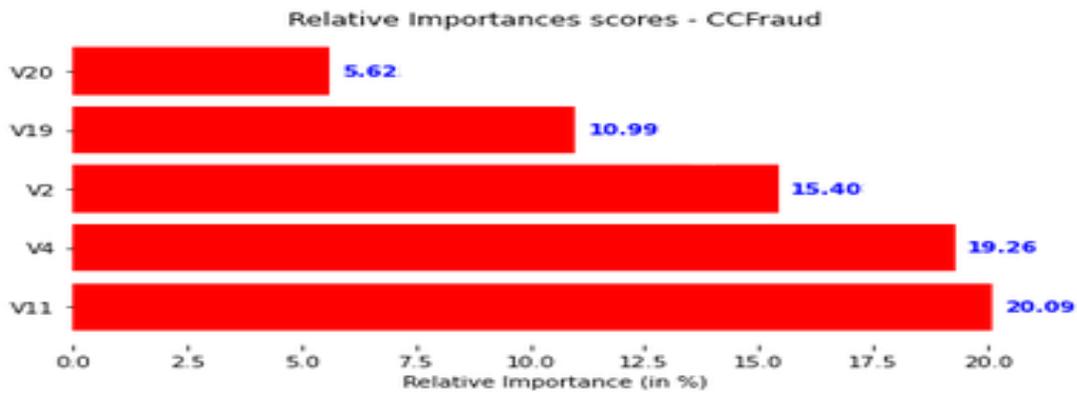

Fig B 14: Contribution of top 5 features using (20, 14, 20, 29) hidden neurons

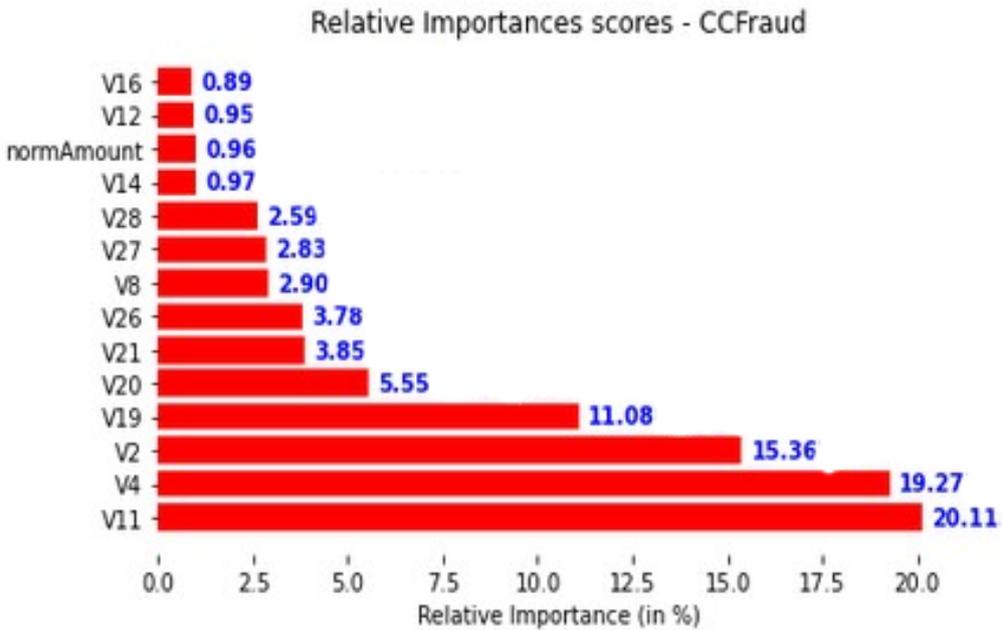

Fig B 15: Contribution of top 14 features using (20, 29) hidden neurons

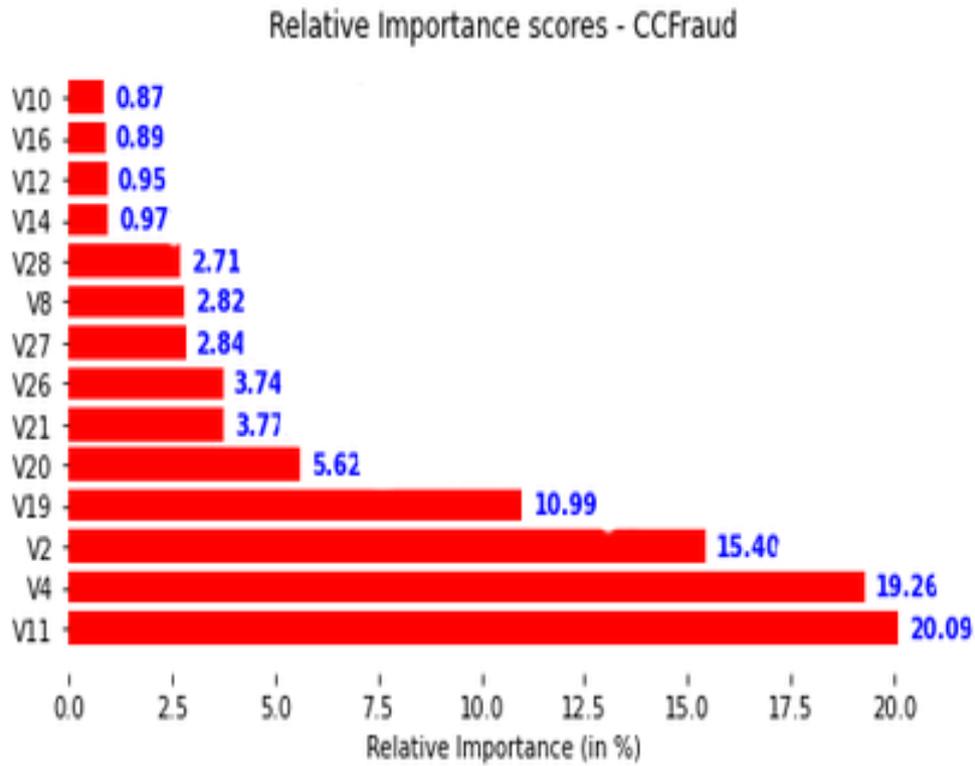

Fig 16: Contribution of top 14 features using (20, 14, 20, 29) hidden neurons

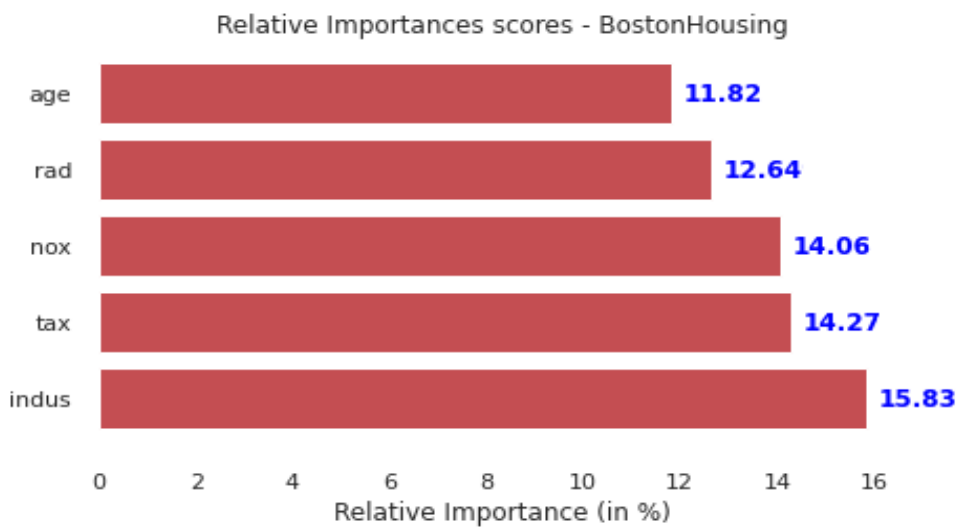

Fig B 17: Contribution of top 5 features using (20, 13) hidden neurons

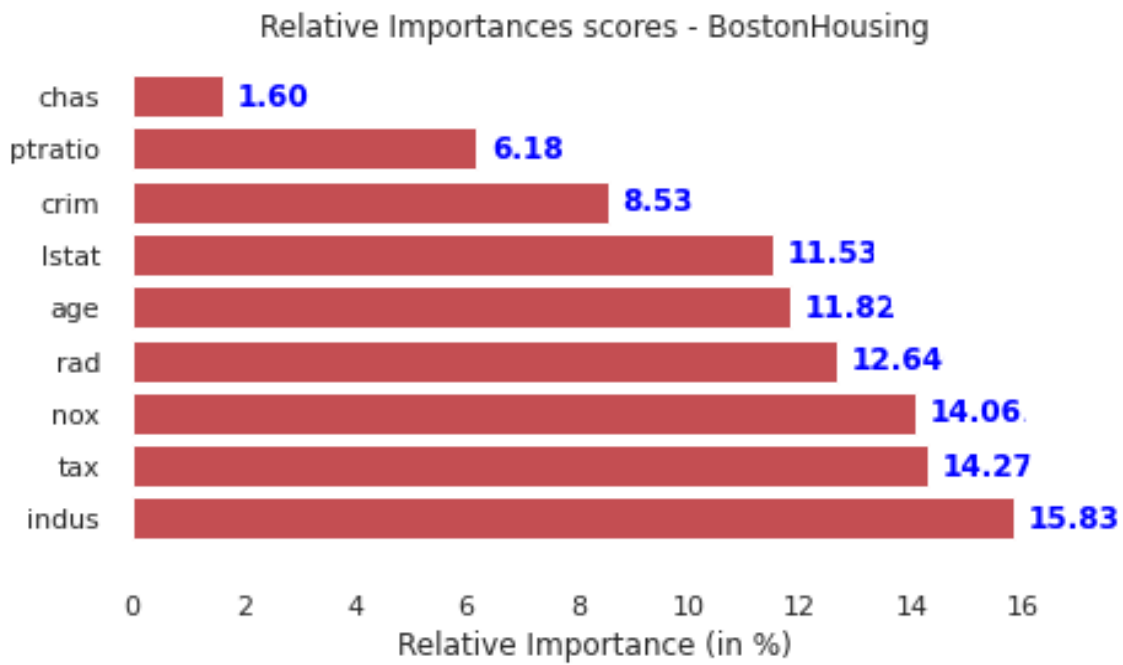

Fig B. 18 Contribution of top 9 features using (20, 13) hidden neurons

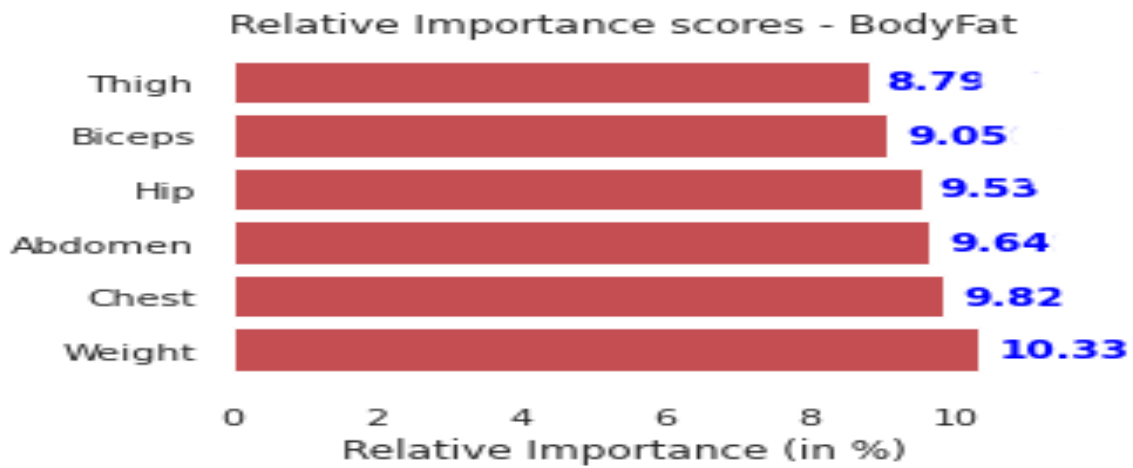

Fig B. 19 Contribution of top 6 features using (10, 12) hidden neurons

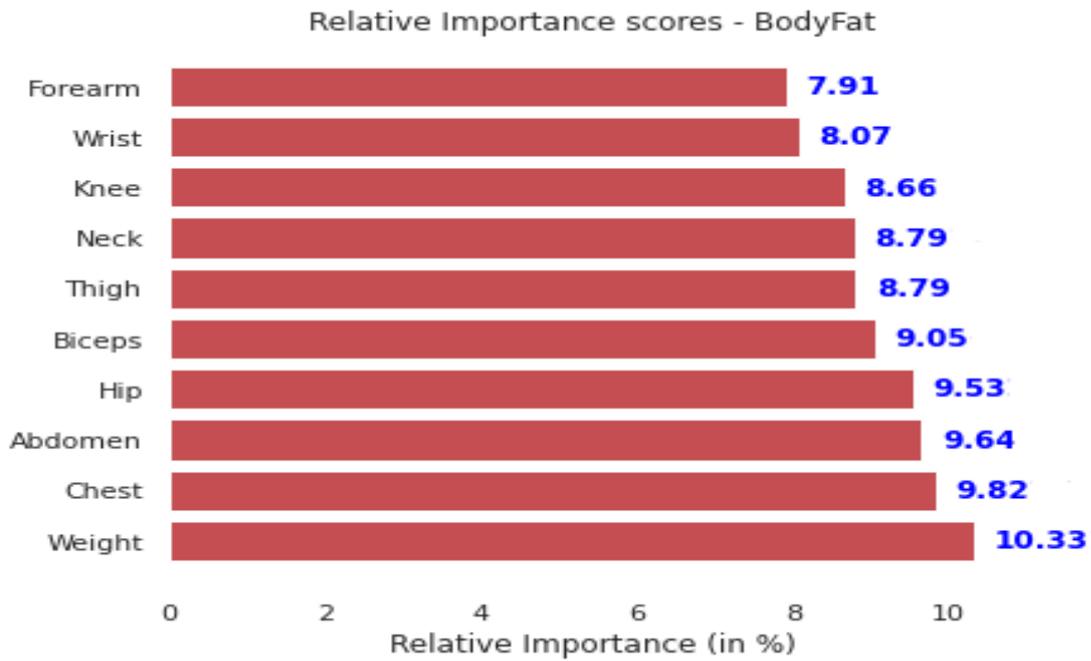

Fig B.20 Contribution of top 10 features using (10, 12) hidden neurons

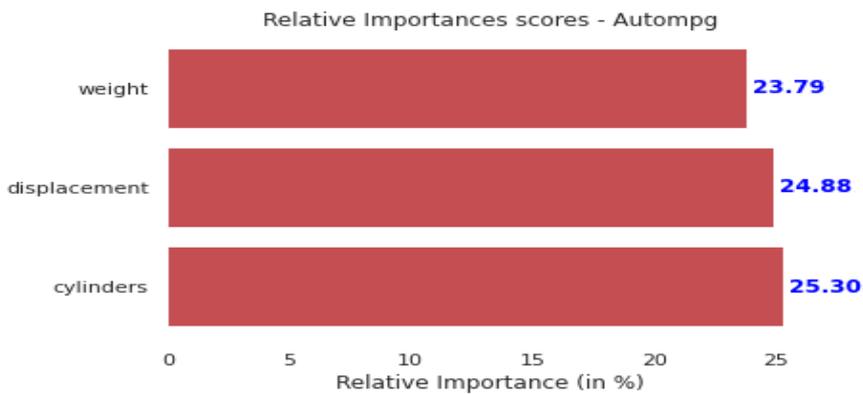

Fig B.21 Contribution of top 3 features using (13, 7) hidden neurons

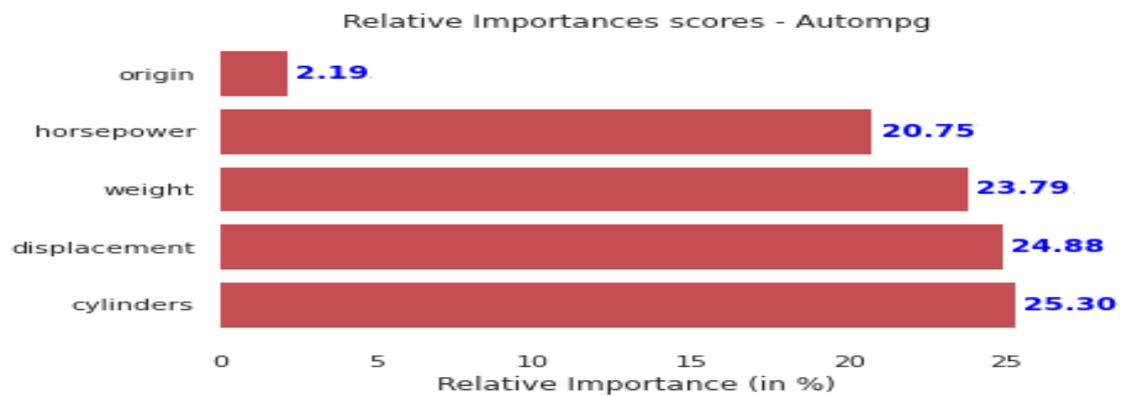

Fig B.22 Contribution of top 5 features using (13, 7) hidden neurons

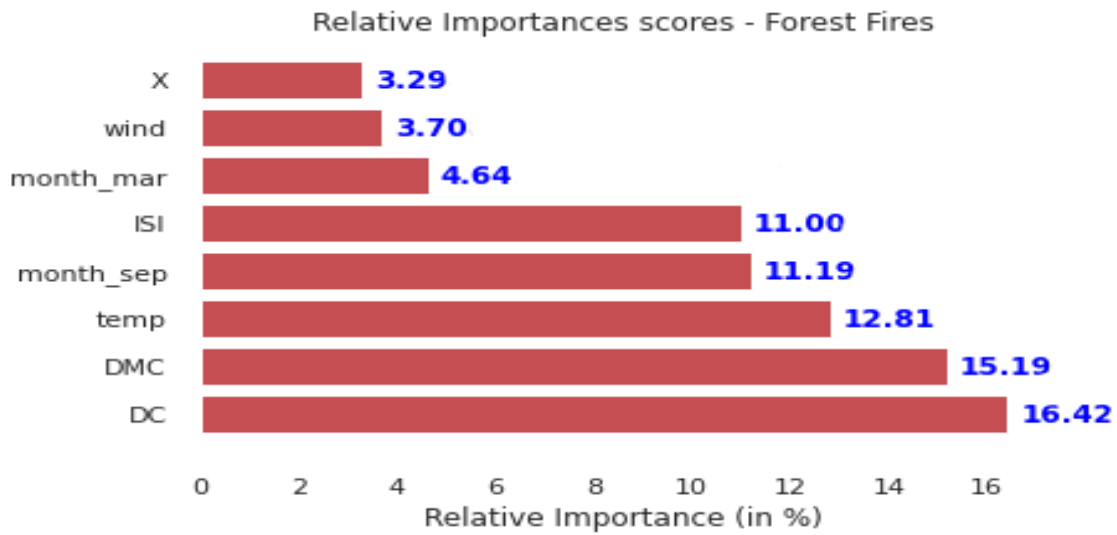

Fig B. 23 Contribution of top 8 features using (15, 21) hidden neurons

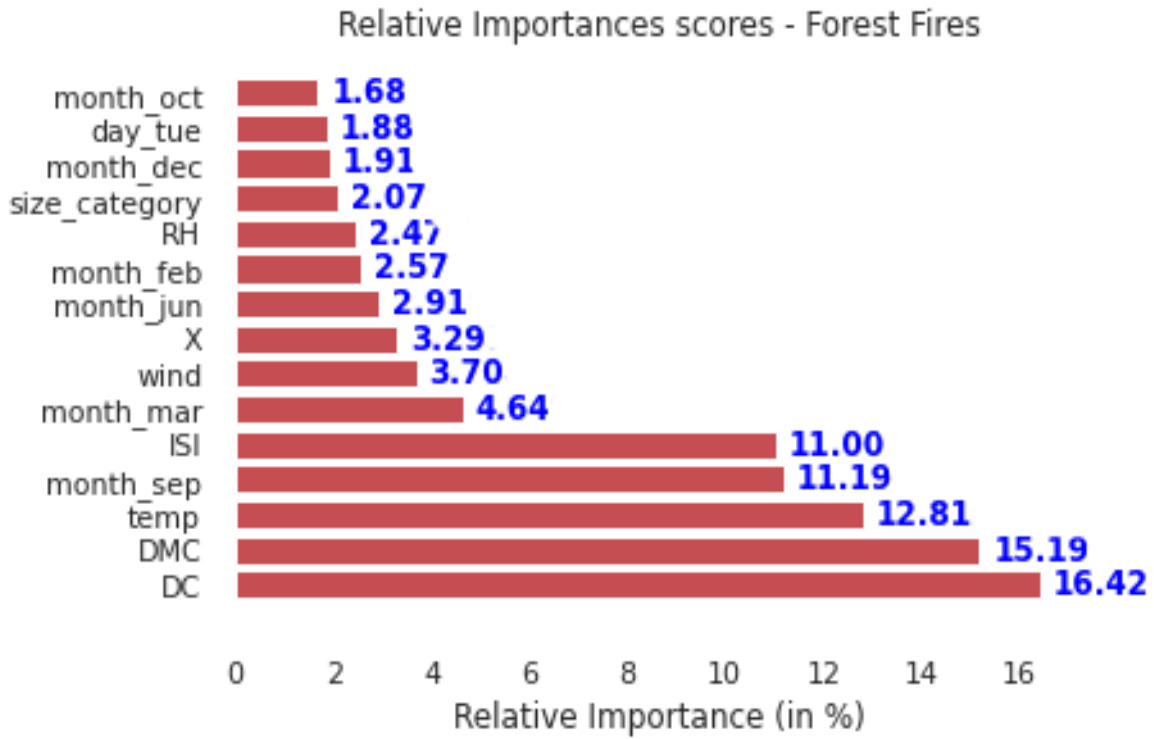

Fig B. 24 Contribution of top 15 features using (15, 21) hidden neurons

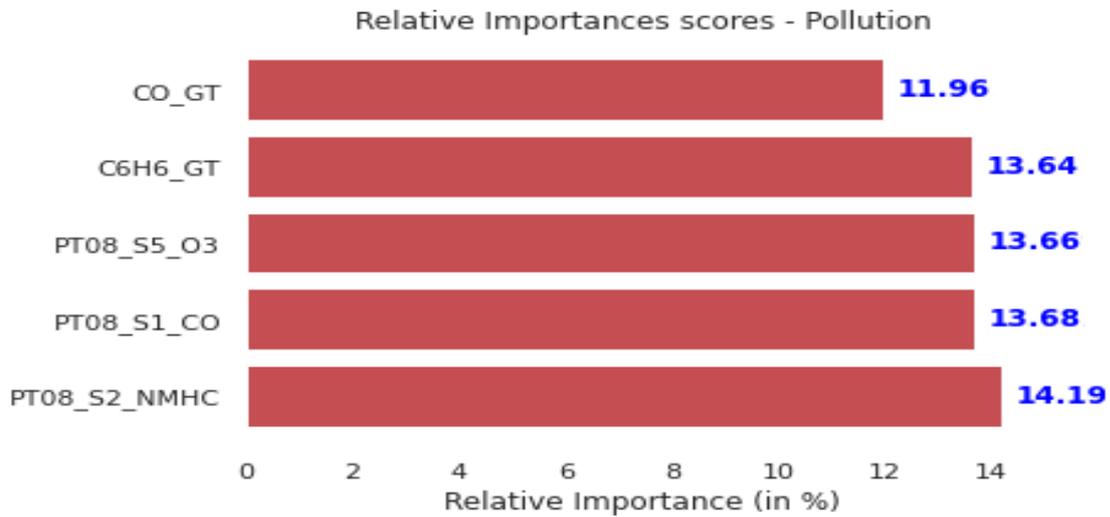

Fig B. 25 Contribution of top 5 features using (15, 10) hidden neurons

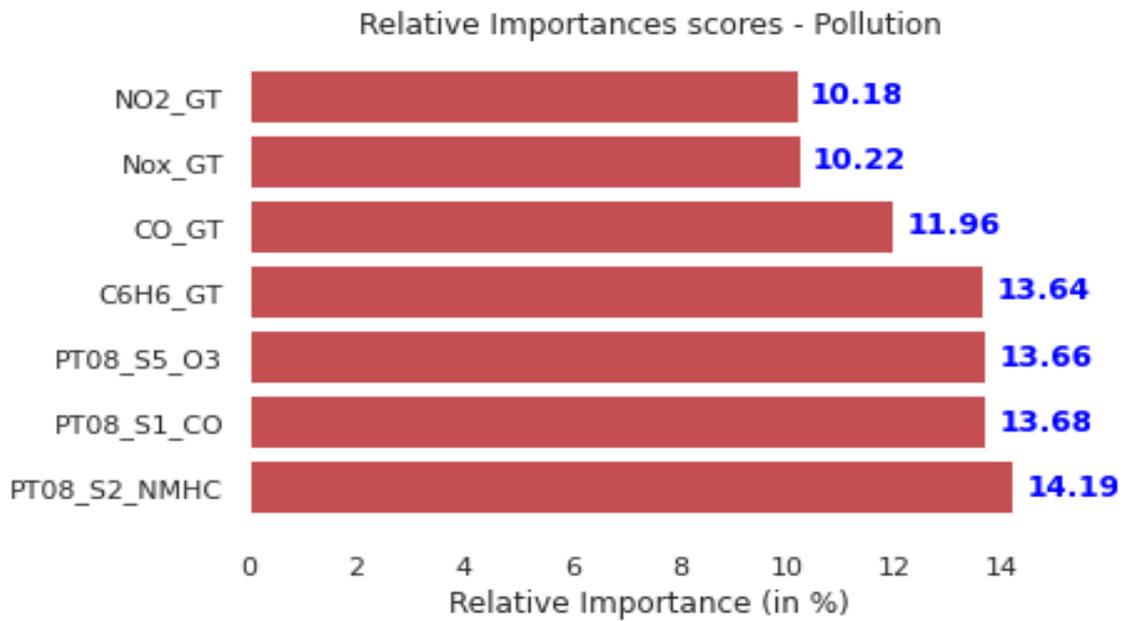

Fig B. 26 Contribution of top 7 features using (15, 10) hidden neurons